\pdfoutput=1

\documentclass[11pt,a4paper]{article}

\usepackage[utf8]{inputenc}

\usepackage[T1]{fontenc}

\DeclareUnicodeCharacter{00A7}{\S}

\DeclareUnicodeCharacter{00B1}{\ensuremath{\pm}}

\DeclareUnicodeCharacter{00B7}{\ensuremath{\cdot}}

\DeclareUnicodeCharacter{00D7}{\ensuremath{\times}}

\DeclareUnicodeCharacter{00E8}{\`e}

\DeclareUnicodeCharacter{00E9}{\'e}

\DeclareUnicodeCharacter{00F1}{\~n}

\DeclareUnicodeCharacter{00F6}{\"o}

\DeclareUnicodeCharacter{00FC}{\"u}

\DeclareUnicodeCharacter{2013}{--}

\DeclareUnicodeCharacter{2014}{---}

\DeclareUnicodeCharacter{2016}{\ensuremath{\|}}

\DeclareUnicodeCharacter{2026}{\ldots}

\DeclareUnicodeCharacter{2192}{\ensuremath{\to}}

\DeclareUnicodeCharacter{21D2}{\ensuremath{\Rightarrow}}

\DeclareUnicodeCharacter{2207}{\ensuremath{\nabla}}

\DeclareUnicodeCharacter{2208}{\ensuremath{\in}}

\DeclareUnicodeCharacter{2212}{\ensuremath{-}}

\DeclareUnicodeCharacter{2218}{\ensuremath{\circ}}

\DeclareUnicodeCharacter{221A}{\ensuremath{\surd}}

\DeclareUnicodeCharacter{221D}{\ensuremath{\propto}}

\DeclareUnicodeCharacter{221E}{\ensuremath{\infty}}

\DeclareUnicodeCharacter{2248}{\ensuremath{\approx}}

\DeclareUnicodeCharacter{2261}{\ensuremath{\equiv}}

\DeclareUnicodeCharacter{2264}{\ensuremath{\leq}}

\DeclareUnicodeCharacter{2265}{\ensuremath{\geq}}

\DeclareUnicodeCharacter{226B}{\ensuremath{\gg}}

\DeclareUnicodeCharacter{2605}{\ensuremath{\star}}

\DeclareUnicodeCharacter{0391}{\ensuremath{\Alpha}}

\DeclareUnicodeCharacter{0392}{\ensuremath{\Beta}}

\DeclareUnicodeCharacter{0393}{\ensuremath{\Gamma}}

\DeclareUnicodeCharacter{0394}{\ensuremath{\Delta}}

\DeclareUnicodeCharacter{0398}{\ensuremath{\Theta}}

\DeclareUnicodeCharacter{039B}{\ensuremath{\Lambda}}

\DeclareUnicodeCharacter{039E}{\ensuremath{\Xi}}

\DeclareUnicodeCharacter{03A0}{\ensuremath{\Pi}}

\DeclareUnicodeCharacter{03A1}{\ensuremath{\Rho}}

\DeclareUnicodeCharacter{03A3}{\ensuremath{\Sigma}}

\DeclareUnicodeCharacter{03A6}{\ensuremath{\Phi}}

\DeclareUnicodeCharacter{03A8}{\ensuremath{\Psi}}

\DeclareUnicodeCharacter{03A9}{\ensuremath{\Omega}}

\DeclareUnicodeCharacter{03B1}{\ensuremath{\alpha}}

\DeclareUnicodeCharacter{03B2}{\ensuremath{\beta}}

\DeclareUnicodeCharacter{03B3}{\ensuremath{\gamma}}

\DeclareUnicodeCharacter{03B4}{\ensuremath{\delta}}

\DeclareUnicodeCharacter{03B5}{\ensuremath{\varepsilon}}

\DeclareUnicodeCharacter{03B6}{\ensuremath{\zeta}}

\DeclareUnicodeCharacter{03B7}{\ensuremath{\eta}}

\DeclareUnicodeCharacter{03B8}{\ensuremath{\theta}}

\DeclareUnicodeCharacter{03BB}{\ensuremath{\lambda}}

\DeclareUnicodeCharacter{03BC}{\ensuremath{\mu}}

\DeclareUnicodeCharacter{03BD}{\ensuremath{\nu}}

\DeclareUnicodeCharacter{03BE}{\ensuremath{\xi}}

\DeclareUnicodeCharacter{03C0}{\ensuremath{\pi}}

\DeclareUnicodeCharacter{03C1}{\ensuremath{\rho}}

\DeclareUnicodeCharacter{03C3}{\ensuremath{\sigma}}

\DeclareUnicodeCharacter{03C4}{\ensuremath{\tau}}

\DeclareUnicodeCharacter{03C6}{\ensuremath{\varphi}}

\DeclareUnicodeCharacter{03C7}{\ensuremath{\chi}}

\DeclareUnicodeCharacter{03C8}{\ensuremath{\psi}}

\DeclareUnicodeCharacter{03C9}{\ensuremath{\omega}}

\usepackage{amsmath,amssymb}

\usepackage{graphicx}

\usepackage{booktabs}

\usepackage{longtable}

\usepackage{hyperref}

\usepackage{url}

\usepackage{enumitem}

\usepackage[margin=1in]{geometry}

\graphicspath{{figures/}}

\title{Deposon: An Auditable, Conservation-Guaranteed, Game-Theoretically Tested Scattering Layer over LLM Reasoning Paths}

\author{Qihao Yuan \\
School of Chemistry and Life Resources, Renmin University of China \\
\texttt{github.com/zeroandcat/Deposon} \\ \thanks{Project version: Deposon v2. The Deposon (author-provided) is a single quasiparticle whose v1 (blocking) and v2 (tunneling) are two limits distinguished only by whether energy dissipates into the infinite-dimensional orthogonal aether, with T+R+A=1 conservation per path.}}

\date{2026-11-03}

\begin{document}

\maketitle

\noindent\textbf{arXiv}: cs.AI \quad \textbf{Repository}: \url{https://github.com/zeroandcat/Deposon} \quad \textbf{Date}: 2026-09-08 \quad \textbf{License}: CC BY 4.0

\vspace{1em}

\hrule

\vspace{1em}

\begin{abstract}

Multi-step LLM reasoning lacks a machine-recheckable ledger: discarded reasoning paths leave no auditable record. We propose the Deposon scattering layer, which binds each node of an LLM-generated concept-decomposition graph to a two-parameter Deposon state; paths undergo three-channel scattering—transmission, reflection, irreversible dissipation—obeying T+R+A=1 for arbitrary parameters, with a maximum per-path energy-audit deviation of 2.2×10$^{-16}$ (machine epsilon). We report all three evidence tiers honestly. On synthetic trap benchmarks the path-filtering gain is closed (pre-registered): unified reaches 100\% versus a decoy-capture baseline at 7\%/10\%. On real benchmarks the layer is indistinguishable from a trivial six-keyword rule filter (GSM8K 0.87 ≥ 0.85, McNemar p=0.5; StrategyQA 0.899 = 0.899); no difference is detected here, so we sharpen the claim to "the differential value lies solely in machine verifiability." Fusion yields a second negative result: convex combinations with a semantic prior never improve (physics 0.484→0.452), and the apparent λ=2 gain is an anti-field artifact; any fusion gain must be nonlinear. Modeling the reverse dynamics as a potential game on the graph, we evidence an auditable scalar's monotonicity and near-gradientness and quantify the empirical coordination ratio (ECR). The three formalized dynamical-equivalence propositions (P1a/P1b/T-P1c) are falsified under the pre-registered kill protocol, and the potential-game claim is downgraded to approximate (cyclic-graph median residual 0.669): only consistency-level evidence survives at the dynamical level. Code: github.com/zeroandcat/Deposon.


\end{abstract}

\section{Introduction}

Chain-of-thought \cite{key01} and its descendants—search-based decomposition \cite{key02} and self-consistency voting \cite{key03}—organize LLM reasoning into explicit intermediate steps and expand them into a space of candidate paths; yet chain-of-thought text need not be faithful to the internal computation \cite{key04}, and faithfulness requires dedicated evaluation \cite{key05}. Three structural difficulties afflict the reliability of multi-step reasoning: superficially related decoys in the problem statement systematically attract search branches; errors in intermediate steps have no intrinsic absorption mechanism and compound along the chain; and the basis for path selection is uninterpretable, so that after the fact one cannot answer "why should this path have been discarded." Whatever the organizational scheme, one structural question remains open: when a path is eliminated, the system cannot produce a recheckable account. Pruning, voting, and heuristic scoring all make elimination decisions, yet the elimination itself leaves no conserved record—how much budget the discarded branch consumed, why it deserved discarding, and whether a third party could recompute the decision step by step have no answer within the framework. Algorithmic auditing research supplies accountability frameworks at the organizational level \cite{key06}; verifiable computation at the cryptographic layer \cite{key07} and the tightening of differential-privacy auditing \cite{key08} demonstrate that verifiability itself can be a first-class design goal. But at the level of LLM reasoning paths, an empty layer sits between post-hoc provenance and cryptographic proof—run-time, per-instance invariants—and we are not aware of prior work occupying this layer.

This paper occupies that empty layer with a physical construction. The vocabulary can be borrowed directly from scattering theory, and we fix it once here: the three-state verdict (transmit / block / tunnel) corresponds to the three scattering channels; dissipation corresponds to a commitment device—once energy condenses into the infinite-dimensional aether it cannot flow back, and the budget of an eliminated path cannot be resurrected; conservation corresponds to the accounting identity—every unit of energy has exactly one destination. The design goals of this project's precursor design phase (preceding the frozen pipeline build) supplied the embryo of this construction: a two-parameter Deposon state (path coupling g\_couple, aether coupling g\_aether), a scattering form inspired by Feshbach resonance, and the axiomatic stipulation "dissipation = irreversible condensation." The route from that definition to the conservation guarantees and auditable representation of §2, and then to the game-theoretic formulation of the reverse dynamics in §3, is two successive tightenings of the same entity; the details are left to those sections.

Writing physical properties into algorithms is not new, but most precedents inject soft constraints: residuals are suppressed by gradients yet can still be violated at inference time. Our construction sits at the opposite pole—conservation holds by construction, independent of training and of tuning, and an auditor needs only double-precision arithmetic to recheck it. This determines the shape of our evidence: the strongest claims (conservation, auditability) rest on no benchmark accuracy whatsoever, while the weakest claim (the benefit of the dissipation channel) is explicitly demoted by our own experiments to "motivation only."

One declaration belongs up front: the output of this research line includes, alongside the mechanism itself, an honest boundary drawn by a set of negative results; the two carry equal weight. Pre-registered controlled experiments show that the scattering layer's apparent accuracy advantage is partly an artifact of benchmark construction (once decoy edges are flattened to equal weight, the attribution of the advantage changes), and that on two real benchmarks the layer is indistinguishable from a trivial rule filter. Such "effect size goes to zero" corrections are nothing to be ashamed of—underclaiming is as harmful as overclaiming \cite{key09}, and disciplined reporting norms \cite{key10,key11} together with systematic records of underspecification effects \cite{key12} are exactly the genre this paper follows. We therefore locate the value proposition in auditable representation and conservation guarantees, and label every claim with one of the three strength tiers.

Our contributions:

\begin{itemize}

  \item \textbf{C1 Auditable representation and the accounting identity (closed (pre-registered))}: the constructive definition of three-channel scattering makes T+R+A=1 hold for arbitrary parameters; the per-path energy-audit residual is 2.2×10$^{-16}$; elimination decisions are attributable node by node (§2).

  \item \textbf{C2 Three-tier placement of evidence strength (closed (pre-registered) / consistency / motivation, side by side)}: the path-filtering gain on synthetic benchmarks, the tie with a rule filter on real benchmarks (E9.5), and the zero-benefit statement for the dissipation channel are reported in parallel, without blending tiers (§2.3–§2.4).

  \item \textbf{C3 An exclusionary conclusion on fusion dilution (closed (pre-registered) on the measured λ settings)}: the convex-combination hybrid never exceeds either single arm on any measured setting; the apparent λ=2 gain is an anti-field artifact; any fusion gain can only come from a nonlinear mechanism (§2.5).

  \item \textbf{C4 Game-theoretic evidence (closed (pre-registered))}: the reverse dynamics admits a scalar (the potential) that can be audited against a benchmark; the empirical coordination ratio (ECR) quantifies "how far from optimum"; the temperature frontier demarcates the audit boundary (§3).

\end{itemize}

Roadmap: §2 gives the representation construction, the conservation identity, per-path auditing, and all three evidence-strength tiers of the static line; §3 gives the game-theoretic evidence chain; §4 collects the honest boundaries; §5 concludes; Appendix A gives the number-traceability table.

\section{Auditable Representation and Conservation Guarantees}

This section presents in a self-contained manner the construction and conservation properties of the scattering layer and reports evidence strength as mechanically evaluated against pre-registered verdicts. §2.1 constructs the representation; §2.2 states the accounting identity and the per-path audit; §2.3 reports the path-filtering gain together with its attribution boundary; §2.4 reports the tie against a rule filter and the resulting relocation of the value proposition; §2.5 reports fusion dilution.

\subsection{Representation: from concept graphs to Deposon states}

Given a natural-language question, an LLM backend first decomposes it into a directed concept graph G=(V,E): nodes include numbers, operations, traps (superficially related but semantically irrelevant decoy candidates), answer candidates, and general concepts; BFS generates the set of candidate paths from the start node to an answer node. The scattering layer binds each node v to a Deposon state characterized by three parameters: a path-coupling strength g\_couple≥0, an aether-coupling strength g\_aether≥0, and a resonance energy E$_{0}$. Parameters are not assigned ad hoc after the fact but bound to node-type semantics: trap nodes take (5.0, 0.0) (strong scattering centers), operation nodes take (0.3, 0.2), all other nodes take (0.05, 0.05), and g\_couple is further multiplied by a small-degree correction. The detuning δ is defined as the difference between the path energy and the node's resonance energy; a Lorentzian factor 1/(1+δ$^{2}$) modulates the effective coupling g\_eff=g\_couple/(1+δ$^{2}$). The scattering formulas are inspired by the Feshbach-resonance form, but we state explicitly: the T/R/A weights below are constructive definitions, and their normalization is a design choice, not a consequence of any S-matrix. In the main experiments every node's resonance energy identically equals its own energy, so δ≡0 and the resonance channel is dormant—all measured effects are driven by the type-wise contrast of (g\_couple, g\_aether). We spell this out so that readers do not overestimate the current role of the resonance mechanism.

The two working modes of the initial definition thereby unify as limiting states of a single entity: at g\_aether=0 energy splits only between transmission and reflection (the blocking state, in which erroneous paths decay by local reflection); at g\_aether≫0 erroneous energy condenses into the aether while correct paths transmit almost losslessly (the tunneling state); at general parameters all three channels are open and behavior interpolates continuously in the ratio η=g\_aether/g\_couple. The content of this unification goes no further than "the limiting behavior of a two-parameter continuous family at its parameter boundaries," and its consistency check is that energy allocation at three parameter points must agree qualitatively with the two limits (verified in the text of §2.2).

\subsection{The accounting identity and per-path auditing}

Define the normalization constant Λ=1+g\_eff+g\_aether. The three-channel energy-allocation weights are T=1/Λ, R=g\_eff/Λ, A=g\_aether/Λ, corresponding to transmission, reflection, and irreversible dissipation into the aether. By construction T+R+A=1 immediately: with the dissipation channel explicitly included, the total energy of system plus environment is strictly conserved. Transmitted energy recurses along a path as E$^{(i)}$=E$^{(i-1)}$T\_i, while cumulative reflection and cumulative dissipation are sums of per-node shares; for any path and any parameters, E$^{(n)}$+E\_refl+E\_diss=E$^{(0)}$ (a one-step induction). The point of this identity is not mathematical difficulty but auditability: every unit of energy either reaches the endpoint, is reflected, or condenses into the aether—exactly one of the three—and a third party can independently recompute this for each path.

The audit is measured, not promised. Across all variants, all 200 synthetic problems, and all candidate paths tested scattering by scattering, the maximum deviation of $|T+R+A−1|$ is 2.220446049250313×10$^{-16}$—exactly the scale of double-precision machine epsilon and far below the 10$^{-6}$ implementation tolerance (closed (pre-registered); \texttt{physics\_\allowbreak\allowbreak {}audit} field of \texttt{results/deposon\_\allowbreak\allowbreak {}v19\_\allowbreak\allowbreak {}benchmark\_\allowbreak\allowbreak {}fixes.json}, passed=true). On the trap benchmark, the per-path average dissipation of the three limiting states is 0 / 3.63 / 0.358 respectively (energy units; \texttt{deposon\_\allowbreak\allowbreak {}benchmark\_\allowbreak\allowbreak {}v1\_\allowbreak\allowbreak3\_\allowbreak\allowbreak {}traps.json → variant\_\allowbreak\allowbreak {}results.{v1\_\allowbreak\allowbreak {}blocking,v2\_\allowbreak\allowbreak {}tunneling,unified}.avg\_\allowbreak\allowbreak {}ether\_\allowbreak\allowbreak {}dissipated}), qualitatively matching theoretical prediction: the blocking state dissipates nothing (closed system), the tunneling state dissipates heavily (open system, erroneous energy condensing in bulk), and the mixed state dissipates moderately. The aether channel is one-directional at the interface level: \texttt{dissipate()} atomically accumulates energy into a monotone counter, and the system exposes no \texttt{recover()}; this corresponds to the physical argument that the premise of Poincaré recurrence fails in an infinite-dimensional orthogonal environment. Honest disclosure: in a finite-dimensional software implementation, "irreversibility" is a combination of an engineering lock and an asymptotic property; the infinite-dimensional approximation cannot be verified numerically and remains an irreducible idealization.

The conservation audit also played a guarantor's role in one error correction. In the frozen pipeline build the high\_couple variant was a pure alias of v1\_blocking (a configuration bug), so its then-reported GSM8K figure of 0.86 was wrong; after a true fix (g\_couple×5 over the whole field, g\_aether=0) an offline rerun gave 0.82, with 4 problems flipped and McNemar versus v1\_blocking p=0.125, not significant (E9.3, \texttt{deposon\_\allowbreak\allowbreak {}v19\_\allowbreak\allowbreak {}benchmark\_\allowbreak\allowbreak {}fixes.json}); on StrategyQA the post-fix prediction vector was bitwise identical to v1\_blocking (p=1.0)—the physical perturbation genuinely occurred but does not alter greedy ranking on shallow 3-step graphs. Both reruns passed the conservation audit at 2.2×10$^{-16}$: the same ledger constrains the physics layer and our own correction process.

\subsection{Path-filtering gain and its attribution boundary}

On two controlled synthetic benchmarks of 100 problems each (seed=42, real LLM-backend decomposition, zero fallback in the final run), the full pipeline (unified variant) reaches 100\% on both the simple set and the trap set, while the same-graph field-free greedy baseline reaches only 7\%/10\% (\texttt{variant\_\allowbreak\allowbreak {}results} of \texttt{deposon\_\allowbreak\allowbreak {}benchmark\_\allowbreak\allowbreak {}v1\_\allowbreak\allowbreak3\_\allowbreak\allowbreak {}simple.json} / \texttt{\_\allowbreak\allowbreak {}traps.json}). Two boundary statements must travel with these numbers. First, the baseline is a decoy-capture baseline: decoy edges are deliberately weighted 0.9 (above the correct operation edges' 0.6) at graph-construction time, and all 93 failed problems select the decoy path; the effect sizes +0.93/+0.90 measure same-graph path-filtering increment on an adversarially weighted graph, not a general capability improvement. Second, a three-way ablation shows that the increment comes from the combination of labels and dynamics: with type labels randomly permuted, trap-set accuracy falls to 17.2\%±6.4\% (± is the sample standard deviation over 5 seeds, ddof=1; t95 half-width 7.9pp), and with uniform parameters on all nodes, trap-set accuracy degenerates to 10\% (numerically coinciding with the field-free baseline; \texttt{deposon\_\allowbreak\allowbreak {}benchmark\_\allowbreak\allowbreak {}v1\_\allowbreak\allowbreak3\_\allowbreak\allowbreak {}labelshuffle.json} → \texttt{uniform\_\allowbreak\allowbreak {}params.accuracy}). The most accurate reading of the scattering layer is therefore a transducer: it converts semantic type labels into auditable energy decisions, and its increment is conditional on label quality.

The picture on real benchmarks is more restrained. On the GSM8K subset (n=100, seed=42): CoT baseline 97.0\%, unified 85.0\%, v1\_blocking 86.0\%, and exact McNemar for unified versus CoT gives p=4.9×10$^{-4}$—CoT is significantly better, and we report the negative conclusion as it stands (\texttt{deposon\_\allowbreak\allowbreak {}benchmark\_\allowbreak\allowbreak {}v1\_\allowbreak\allowbreak4\_\allowbreak\allowbreak {}gsm8k.json}). The 95\% confidence interval for the 85\% versus 97\% difference is [−20.5pp, −4.1pp] (unpaired Newcombe hybrid interval, same caliber as §2.4; \texttt{deposon\_\allowbreak\allowbreak {}v22\_\allowbreak\allowbreak {}e95ci.json} → \texttt{unified\_\allowbreak\allowbreak {}vs\_\allowbreak\allowbreak {}cot}), and the binomial CIs are computed by the Newcombe-Wilson method. On StrategyQA (n=99): unified 89.9\% versus CoT 92.9\%, p=0.549, no significant difference (\texttt{deposon\_\allowbreak\allowbreak {}benchmark\_\allowbreak\allowbreak {}v1\_\allowbreak\allowbreak4\_\allowbreak\allowbreak {}strategyqa.json}); the three arms v1\_blocking, high\_couple, and unified all score 89.9\%—the constraint layer produced no differential action whatsoever on this task, a "constraint-layer inertia" that we record honestly. Read together, the two real benchmarks reveal a task-dependent constraint–fidelity trade-off: on GSM8K's clean long chains the cost of information loss dominates, while on StrategyQA's short chains of implicit reasoning the constraint layer ties CoT; the layer's cost varies with chain length and decomposition fidelity. The equal-weight decoy control (E9.4, pre-registered) further severs misattribution: after flattening every edge weight of the same cached concept graphs to 0.7, the unified advantage does not vanish (GSM8K 0.85 versus no\_deposon 0.04, p=1.7×10$^{-23}$; StrategyQA 0.899 versus 0.202, p=7.5×10$^{-15}$; E9.4 field of \texttt{deposon\_\allowbreak\allowbreak {}v19\_\allowbreak\allowbreak {}benchmark\_\allowbreak\allowbreak {}fixes.json}), but its source is located in BFS's shortest-path-first ordering and the type='trap' labels the physical layer gets for free—not in the scattering mechanism itself. The apparent effect size of unified versus no\_deposon under the main protocol is therefore an unattributable number under structural bias, and this paper does not cite it as anti-capture value.

\subsection{The E9.5 tie: differential value lies only in machine verifiability}

The sharpest control is a trivial baseline. We construct a purely deterministic rule filter: after the same greedy path generation, discard any path passing through a node whose label hits the six-keyword list {trap, dead, end, impossible, guess, wrong}—reading only label strings, never type metadata, and involving no scattering mechanism at all. The result (E9.5, mechanically judged after pre-registration): on GSM8K the rule filter scores 0.87 ≥ unified 0.85 (McNemar b=0, c=2, p=0.5); on StrategyQA 0.899 = 0.899 (b=0, c=0, p=1.0). On two real benchmarks, the three-channel scattering pipeline is indistinguishable from a six-keyword filter on the accuracy dimension. The power-honest statement: no difference is detected at this sample size—the GSM8K difference is −2pp (95\% CI [−11.8pp, +7.8pp], Newcombe-Wilson) and the StrategyQA difference is 0pp (95\% CI [−8.8pp, +8.8pp])—both are unpaired Newcombe hybrid intervals (conservative bound; paired discordant pairs are only 0/2 and 0/0, so paired intervals degenerate; computation artifact \texttt{deposon\_\allowbreak\allowbreak {}v22\_\allowbreak\allowbreak {}e95ci.json}); with only 2/0 discordant pairs, the detectable-difference threshold is about ±10pp, so smaller increments cannot be excluded. The claim is therefore sharpened to "the differential value lies solely in machine verifiability," supported by the E9.4 attribution mechanism rather than by the strong form of this test.

The mechanism behind the tie is no mystery: the scattering parameters are driven by node-type labels, and the rule filter reads the string form of those same labels—same information source, same discriminative power. The scattering layer does not squeeze more discriminative information out of the labels; what it does is transcribe the same information into a representation that carries a conservation ledger.

This negative result is not a footnote of failure but the basis for relocating the value proposition; the paper executes the pre-registered kill rules and claims nothing beyond the data \cite{key09}. What the rule filter lacks and the scattering layer alone possesses is: a per-path T+R+A=1 conservation ledger at arbitrary parameters (machine-precision residual), a recheckable energy record left by every block, and elimination decisions attributable node by node to concrete (T,R,A) shares. In other words, the differential value of the scattering layer lies only in machine verifiability, not in filtering performance. A parallel zero-benefit statement: the irreversible dissipation channel has never outperformed its disabled counterpart on any task measured so far (v1\_blocking and unified tie on the synthetic benchmarks; GSM8K 86.0\% ≥ 85.0\%; all three arms tie at 89.9\% on StrategyQA); its benefit claim currently has only theoretical motivation—erroneous energy cannot be resurrected within a single run—and validation requires iterative re-search scenarios, which we list as future work. The conditional value of LLM semantic signals on graph tasks aligns with existing chains of evidence \cite{key13,key14,key15}; that concept-map evaluation is protocol-sensitive is a thirty-year-old lesson \cite{key16,key17}, of which E9.4/E9.5 are contemporary instances.

\subsection{Fusion dilution: convex combination improves on no setting; gains can only come from nonlinear fusion}

Beyond the scattering layer, the same concept-graph completion task admits a semantic-prior arm (a labels-only LLM prior with zero leakage: the prompt contains only node labels). A natural hypothesis holds that the two are complementary—the field handles structure, the prior handles semantics—so a convex combination hybrid=λ·field+(1−λ)·prior should take the best of both. A pre-registered scan refutes the weakest operational implication of this hypothesis. In the four-λ single-graph scan with λ∈{0.25,0.5,1,2} all four settings are identical (dimensional saturation already at λ=0.25): named Hits@3 is constant at 0.294 and \texttt{any\_\allowbreak\allowbreak {}lambda\_\allowbreak\allowbreak {}pass=false} (\texttt{success\_\allowbreak\allowbreak {}evaluation} of \texttt{deposon\_\allowbreak\allowbreak {}v16\_\allowbreak\allowbreak {}llm\_\allowbreak\allowbreak {}prior.json}). In the family-L four-graph all-candidates protocol at λ=0.5 (\texttt{deposon\_\allowbreak\allowbreak {}v20\_\allowbreak\allowbreak {}crossval.json}, \texttt{hybrid\_\allowbreak\allowbreak {}lambda\_\allowbreak\allowbreak {}convex=0.5}): physics 0.484→0.452, historical 0.783→0.739, and the other two graphs flat—the hybrid never exceeds the prior-only arm on any measured setting; the field only dilutes a genuine semantic prior.

A further ablation exposes how an apparent gain can be faked. The λ=2 setting (field coefficient −1, prior blank rows effectively reverse-sorted by field score) once produced an apparent "fusion gain" of named Hits@3=0.471 under a normalized variant (the \texttt{hybrid\_\allowbreak\allowbreak {}norm@2.0} arm of \texttt{deposon\_\allowbreak\allowbreak {}v17\_\allowbreak\allowbreak {}fusion\_\allowbreak\allowbreak {}fix.json}); the E9.6 null ablation, which fixes the endpoints and shuffles only the confidence values, yields named scores exactly equal to the real prior (0.1176=0.1176, with the random-edge null hypothesis scoring 0 on all 5 runs; \texttt{E9\_\allowbreak\allowbreak6c\_\allowbreak\allowbreak {}lambda2\_\allowbreak\allowbreak {}null\_\allowbreak\allowbreak {}ablation} of \texttt{deposon\_\allowbreak\allowbreak {}v19\_\allowbreak\allowbreak {}quickwins.json})—that hit is not attributable to semantic confidence, only to endpoint position and the anti-field artifact, and the per-edge annotations are on record. Together the two lines of evidence form an exclusionary conclusion (closed (pre-registered) on the measured λ settings; we do not extrapolate it to a theorem over the whole λ space): the convex-combination channel is closed, and if field–prior gains exist at all, they can only come from nonlinear fusion mechanisms. The GNN literature already answers systematically how value domains divide between structural and semantic signals—the effective domain of structural bias is determined by graph properties, and label signals take over on heterophilous graphs \cite{key18,key19}. Our dilution conclusion points the same way and pushes the demarcation to the sharper setting of "extreme signal pairs that are mutually blind at the mechanism level."

\section{The Game-Theoretic Formulation: Potential, the Empirical Coordination Ratio (ECR), and the Audit Boundary}

The conservation ledger of §2 is verified only statically: it proves that the energy of each scattering event goes where it should, but it does not answer the dynamical question—does the reverse evolution of the field, taken as a whole process, possess a scalar against which "where to step next, how far one has come, and how far one remains from optimum" can all be audited on the same ledger? This section models the reverse dynamics as a potential game on the graph and answers under a single pre-registered protocol on 22 controlled concept graphs. We fix the reporting convention first: the positive narrative of this section adopts the consistency register—the evidence is consistent with a potential-game reading, i.e., consistency evidence rather than formal proof; wherever a pre-registered verdict closed (GT-5b, GT-6) or a formalization kill-test closed (GT\_FORMAL, T-P1c), we label it separately as "closed (pre-registered)" and never blend tiers. The section gives its operational definitions in a self-contained manner and relies on no other document.

Task and field are defined as follows. The task is concept-graph completion: leave-one-out prediction over each gold edge, candidates are all graph nodes, and the metric is named Hits@3. For a leave-one-out task (s,t), the scattering layer defines a field-guided energy on the graph's adjacency weight matrix W (row-stochastic): E(s,t)=−log(Σ\_p Π\_e t\_e)+λ\_smooth·Σ W$^{2}$\_ij, where the first term is the negative log of the aggregate transmissivity over all s→t paths and the second is a smoothness regularizer. The reverse process performs simplex natural-gradient descent on masked positions under an annealing schedule (β linear, 50 steps, lr=0.1), formalized as the update rule w\_{t+1}∝(1−lr)·w\_t∘exp(lr·w\_t∘∇Φ), projected back onto the simplex at each step; upon termination the distribution over the masked row ranks the candidate targets, yielding field scores. The deterministic mean-field reverse (field\_mean) takes the mean of a Dirichlet starting distribution and involves no sampling noise anywhere; the control arm introduces Dirichlet sampling at the start, whose concentration parameter α plays the role of an effective temperature. The corpus is designed along two families, "structural negation × genuine semantics": family S (16 graphs, synthetic placeholder labels decoupled from structure) negates the strong claim "field = universal skeleton detector"; family L (6 graphs, genuine domain concept labels, LLM-generated, 30–45 node DAGs) carries the semantic-arm tests.

\subsection{The model: a potential game on a finite graph}

The modeling quadruple is fixed in one pass: each leave-one-out prediction edge is a player; strategies are candidate target nodes; utilities are field scores; and the negative physical energy Φ=−E is the potential-function candidate. Classical existence conditions for potential games come from Monderer \& Shapley \cite{key20} and Rosenthal's congestion games \cite{key21}; Sandholm's population best-response dynamics \cite{key22} prove that deterministic best-response dynamics ascend the potential gradient in the population limit—the nearest theoretical anchor for reading "mean-field deterministic reverse as noiseless best-response dynamics"; Candogan et al.'s flow decomposition of games \cite{key23} provides the operative analogy for "field = graph-flow component" (we decompose the edge-utility vector and do not integrate along trajectories). The decomposition "each task is an independent player" is a modeling choice, not a unique one; Φ=−E has analytic grounding—the reverse annealing gradient is exactly −∇E. One point must be stated plainly: the leave-one-out tasks are mutually independent (each has its own (s,t) and its own energy function), so the existence of an exact potential at the game level is trivial—Φ=Σu\_i already proves it, and no test is needed. GT-6's contribution is therefore not nontrivial evidence for game-potential existence but the near-gradient property of the per-task energy function as a scalar ledger: in the edge-utility vector space (Euclidean inner product), the edge-utility vector F is projected onto the gradient subspace, p=pinv(B)F, with residual ratio r=‖F−Bp‖/‖F‖; r≈0 means the scalar ledger explains the edge utilities almost completely (§3.2).

The model lends three concepts new semantics. First, dissipation is an endogenous commitment device: at g\_a$>$0 energy condenses into the aether and cannot flow back, the budget of an eliminated branch cannot be resurrected, and no player can undo an elimination by rewriting the books after the fact—commitment is not an external rule but a property of the dynamics itself. Second, auditability upgrades from stepwise compliance checking to a scalar ledger: if a potential exists, the entire trajectory can be audited as the stepwise ascent of a scalar curve Φ(t) (equivalently the stepwise descent of E(t)); a third party need not recompute every energy component at every step, only recheck whether one scalar sequence is monotone—the dynamical-layer counterpart of the conservation ledger. Third, "how far from the coordination optimum" becomes computable: we define the empirical coordination ratio (ECR) to quantify the gap between self-interested dynamics and the field benchmark. ECR is the distribution-level operational counterpart of the classical worst-case PoA (the Koutsoupias \& Papadimitriou sense \cite{key24}; the affine-congestion PoA=4/3 bound of Roughgarden \& Tardos \cite{key25}), not the same metric; to our knowledge, such a distribution-level operational counterpart has rarely been reported systematically (distribution-level reporting has precedent in \cite{key26}, and classical equilibrium-efficiency analysis in \cite{key27}). Scope qualifications are in §3.2.

\subsection{Decisive evidence: monotonicity, near-gradientness, and quantification of the auditable scalar (closed (pre-registered) tier)}

Two pre-registered verdicts closed and one closed under restrictions; together they answer "does the scalar exist, and to what quantitative precision."

\textbf{GT-5b potential-trajectory monotonicity (closed (pre-registered), narrowed claim)}: the Φ trajectory of mean-field reverse is monotone non-decreasing on 22/22 graphs, a monotonicity rate of 100\%, above the pre-registered line of 80\%; the kill line—any single triggering graph kills—did not trigger, and the verdict supports\_narrowed\_monotonicity closed (\texttt{deposon\_\allowbreak\allowbreak {}v20\_\allowbreak\allowbreak {}gt5b.json} → \texttt{per\_\allowbreak\allowbreak {}graph\_\allowbreak\allowbreak {}summary.*.meanfield\_\allowbreak\allowbreak {}monotone\_\allowbreak\allowbreak {}rate=1.0}). The claim is the narrowed version: the original GT-5 endpoint condition failed (the noise arm's terminal Φ overtook on 3/4 graphs, judged inconclusive; see §3.5), the monotonicity claim closed independently under a new pre-registration, and the endpoint reversal remains on record, unrewritten. The scope qualification must travel along: 22/22 monotonicity is a frozen fact under the "single-edge leave-out mask × corpus graphs" protocol and is not extrapolated into better-response evidence—under general mask structures that dynamics has already been killed (§3.3).

\textbf{GT-6 non-potential residual (closed (pre-registered), near-gradientness register)}: with the residual ratio r as defined in §3.1 (projection of the edge-utility vector F onto the gradient subspace; cited here, not restated), the median non-potential residual is 1.594×10$^{-29}$, far below the pre-registered line of 0.10; the verdict potential\_game\_explanation\_complete closed (\texttt{deposon\_\allowbreak\allowbreak {}v20\_\allowbreak\allowbreak {}gt6.json} → \texttt{verdict.median\_\allowbreak\allowbreak {}residual\_\allowbreak\allowbreak {}ratio}). Relocation note (see §3.1): since game-level exact-potential existence is trivial under independent tasks (Φ=Σu\_i suffices), what GT-6 tests is not that, but the near-gradient property of the per-task energy function as a scalar ledger. Two honest disclosures. First, three graphs with cyclic structures exceed the line—S4=0.148, L\_algorithm\_process=0.136, S5=0.121—on which the potential explanation is only approximate. Second, the numerical-precision qualification: the median residual sits at the scale of double-precision floating-point underflow and reflects the structural fact that "the edge-utility vector lies almost entirely within the gradient subspace"; it should not be read as a physical quantity by its absolute magnitude, and the 0.12–0.15 residuals of the three exceptional graphs are the informative non-potential components. This decomposition is the operative analog of Candogan's flow decomposition \cite{key23}.

\textbf{GT-4 empirical coordination ratio ECR (restricted-consistent, scope qualification attached)}: the operational definition is ECR=field\_mean/max(self-interested arm), with self-interested arm set {random, degree} (the pre-registered llm\_prior arm is unavailable on family S; the denominator can only be smaller and the ECR only larger—disclosed as-is). The main report covers all 17 finite-valued graphs: median ECR=1.333 $>$ the pre-registered line 1.2; the family-S subset of 13 finite-valued graphs (median 1.5) is listed as a supplementary register; the 3 graphs with ECR=∞ are, per the standing handling rule, counted separately and excluded from the median (self-interested arm named=0 while field$>$0). GT-4 covers 20/22 graphs: L\_geography\_world and L\_project\_management were not evaluated in the frozen run (arm data absent)—neither finite nor ∞; disclosed as-is, the median caliber is unaffected. Two qualifications. First, ECR is the distribution-level operational counterpart of the classical worst-case-NE/social-optimum ratio (PoA), not the same metric: the affine/separable cost-structure premises are not closed, we report at the distribution level only, and we draw no numerical juxtaposition against 4/3 or any other classical worst-case bound—with different metrics, numerical proximity is not evidence. Second, all ECR$<$1 cases are disclosed in parallel: two in family L (L\_historical\_causality 0.5, L\_physics\_concepts 0.75) and, elsewhere in family S, S2\_n45=0.5—the field is negatively coordinating in the semantic domain, which is self-consistent with the division-of-labor boundary of §3.4: the field creates coordination value only in the structural domain, the applicable domain of the potential-game reading coincides with that boundary, and the theory does not contradict itself. Frozen-convention note: the field names in the verdict JSON (\texttt{GT4\_\allowbreak\allowbreak {}price\_\allowbreak\allowbreak {}of\_\allowbreak\allowbreak {}anarchy}, \texttt{field\_\allowbreak\allowbreak {}coordination\_\allowbreak\allowbreak {}value\_\allowbreak\allowbreak {}supported}) are frozen pre-run conventions and are not renamed; the text uses ECR throughout.

\begin{figure}[t]

\centering

\includegraphics[width=0.92\linewidth]{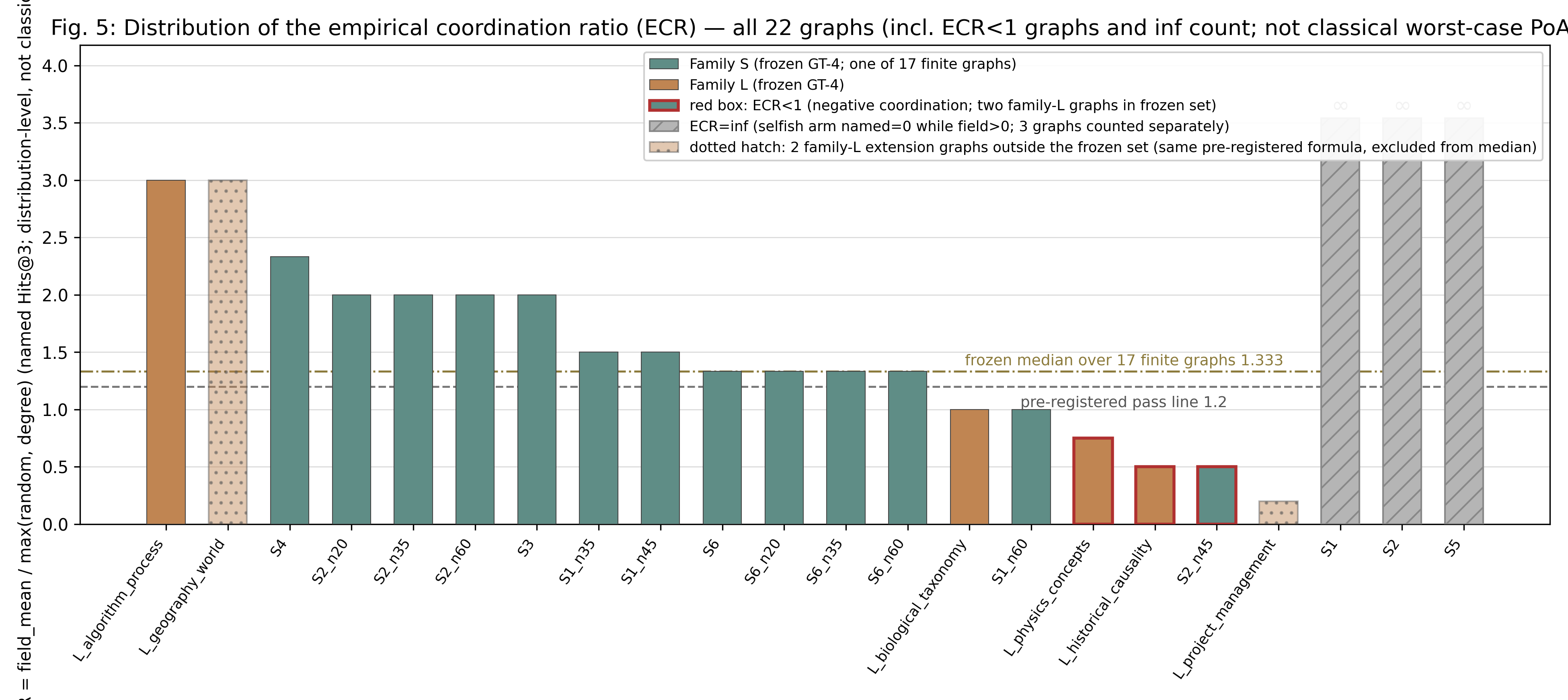}

\caption{Distribution-level empirical coordination ratio (ECR) across all graphs (bars include the red-boxed ECR$<$1 graphs and the separately counted ∞ cases; median 1.3333 and the pass line 1.2 are both read from the frozen verdict fields)}

\end{figure}

(Note on figure order: Figure 5 is first cited here, ahead of Figures 1–4 in §3.4—a forward cross-reference; figure numbers are kept identical to the frozen figure-file names and are not renumbered.)

Taken together: each step of the evolution is auditable as potential ascent (GT-5b + GT-6, closed (pre-registered) under their respective protocols), and the distance of self-interested dynamics from the coordination optimum is computable (GT-4, restricted-consistent). The empirical case for the monotonicity and near-gradientness of the auditable scalar now stands.

\subsection{Formalization kill-tests: all three tiers of dynamical equivalence falsified (closed (pre-registered), systematically sampled graph families × exhaustive-state protocol)}

The model of §3.1 contains an interpretive correspondence: deterministic mean-field reverse = noiseless best-response dynamics. The strongest formalized version of this correspondence is systematically refuted by the GT\_FORMAL kill-test (\texttt{deposon\_\allowbreak\allowbreak {}v21\_\allowbreak\allowbreak {}gtformal.json}, seed=210021; the verdict function was committed as a pure function before any run): 61 systematically sampled small graphs with n≤8 (five families: chain/star/tree/random DAG/cyclic) × the 338 single-node all-candidate mask tasks on them × 20 mean-field steps = a full enumeration of 6760 kill states—i.e., the exhaustion operates at the task/state level, while the graph families are systematically sampled. The strength tier is "closed (pre-registered) (systematically sampled graph families × exhaustive-state protocol)"—what is closed (pre-registered) is the negative fact itself, that each strong formulation fails under this protocol; the kill is the answer, and the strength is not rounded upward. The kill logic does not depend on enumerative completeness: one counterexample kills a universal claim, and 10/338 violations are more than sufficient.

\begin{itemize}

  \item \textbf{P1a strong dynamical equivalence: falsified.} max‖T−BR‖∞=0.8569; an lr scan shows the directional deviation 1−cos≈0.91 does not vanish with lr—an O(1) deviation rather than an O(lr$^{2}$) discretization error, not a step-size problem.

  \item \textbf{T-P1b better-response version: killed.} min directional cosine −1.0, min ΔΦ=−1.2424×10$^{-2}$ (kill line −1e−9), with 10/338 tasks in violation, all on support graphs with non-empty cycle space. The mechanism in one sentence: the fixed point of the composite operator T=Π∘C∘M is not a row-constrained maximizer of Φ; the trajectory keeps advancing past the peak (overshoot), and ΔΦ$<$0 beyond the maximum.

  \item \textbf{P1c residual entropy-regularization characterization: killed (T-P1c, \texttt{deposon\_\allowbreak\allowbreak {}v22\_\allowbreak\allowbreak {}p1c.json}, same sampled-family/exhaustive-state protocol, verdict function pre-registered before any run).} The residual claim was "the first-order update direction = the mirror-ascent direction of the entropy-regularized potential Φ\_τ=Φ+τH" (at τ$>$0 the entropy term might repair overshoot). Two kill lines, both struck: the strong form requires min directional cosine ≥0.999 under a globally uniform τ—an 81-point grid over τ∈[0,4] fails everywhere, and at the best global τ the min cos=−1.0; the weak form allows τ to be chosen per state (τ*) and requires min cos≥0.99—min cos=−1.0, i.e., overshoot states exist whose direction is strictly opposite to every entropy-regularized mirror direction; the median per-state τ* is 0 (85.98\% of states have τ*=0), so the entropy term has no reparative power. The verdict function is locked by 11 tests, and a same-seed rerun is byte-identical.

\end{itemize}

With all three tiers—P1a strong form, P1b better-response, P1c first-order entropy-regularized characterization—killed, the game-theoretic main line retains no residual claim of dynamical equivalence. The protocol relationship needs clarifying: GT-5b's 22/22 monotonicity (single-edge leave-out mask × corpus graphs) is unaffected and remains a closed frozen fact; the present test proves, under the strengthened protocol "all-candidate mask × systematically sampled small graphs," that under general mask structures the dynamics is not better-response. Both protocols stand on record; neither cancels the other.

\textbf{Pre-registration time anchors.} The verdict pure functions and SPECs were frozen before any run, with SHA-256 anchors (first 12 hex digits): GT\_FORMALIZATION\_v1.md = aeefb8ef6972; run\_v21\_gtformal.py = 9bbe43f41fa8; run\_v22\_p1c.py = 6e9673205dc0; SPEC\_GT2B = 68a5b08ef007; SPEC\_GT8C = 6b09de9911c0. Anyone can verify the frozen versions against these anchors and mechanically rerun every verdict.

\textbf{P2 potential completeness: downgraded to approximate potential game.} Acyclic (undirected-forest) support graphs have residual r≤5.9×10$^{-16}$ (numerical zero); support graphs containing cycle space have median r=0.669, and the fraction with r$>$0.30 is 0.924 $>$ the pre-registered 1/3 downgrade line, triggering the downgrade verdict downgraded\_to\_approximate\_potential\_game—same direction as the three cyclic-graph exceptions of GT-6 and of higher magnitude (all-candidate-mask support graphs are denser).

\textbf{P3 dissipation channel: dead in both directions, but yielding the first mechanistic premise evidence.} "Dissipation is merely a reparameterization of payoffs" is killed (max$|r(0.1)−r(0)|$=0.1585 $>$ 1e−9); "the equilibrium set is invariant" is likewise killed (maximum difference of best-response fixed points across three g\_a settings: 0.8442)—dissipation is no pure reparameterization; it genuinely moves equilibrium positions. An incidental finding changed the evidential landscape of the dissipation channel: in 58/338 tasks with g\_a=0 and cycles in the support, ρ(G)=1 makes (I−G) singular and the closed-form dynamics diverges; the spectral condition for convergence, ρ(G)$<$1, is guaranteed precisely by dissipation g\_a$>$0—this is a built-in property of the g\_a$>$0 construction and is archived as mechanistic premise evidence (previously only the negative "no rent" register existed; see §4).

\subsection{Demarcation and boundary regularities: where the auditable advantage holds (directional evidence, not upgraded)}

Once the auditable scalar exists, the next question is "on which graphs does it hold." This subsection states the demarcation regularities with all their qualifications, positioned as observational regularities (directional evidence), not as an established discriminator contribution.

\textbf{Contraction of the validity domain (pre-registered verdict)}: the headline benchmark H-A1 (field\_mean $>$ random) is killed—a 22-graph sign test gives 16+/4−/2 ties, p=0.0118, passing Holm, but the pre-registered kill line is a disjunctive rule (non-significance or ≥3 reversed graphs), and with 4 reversed graphs (L\_historical\_causality, L\_physics\_concepts, L\_project\_management, S2\_n45) the kill line triggers and the headline claim dies. The surviving statement is H-A2 (field\_mean $>$ degree), robust across protocols: 22 graphs, 19+/1−/2 ties, p=4.0×10$^{-5}$, Holm-passed; on the 20-graph subset, Wilcoxon (p=0.0031, $|r|$=0.83, with r=Z/√N) and a paired t-test (p$<$0.0001, d=2.05, Cohen's dz) confirm a large effect; since Hits@3 is a bounded discrete quantity, Wilcoxon is the primary criterion and the paired t-test is reported only as a reference. Multiplicity statement: the Holm correction is applied only within the H-A1/H-A2 family; each GT experiment was independently pre-registered and independently judged, with no pooled correction across families—stated as-is. The field's validity domain thereby contracts to "robust advantage over trivial structural baselines + local advantage on high-hub graphs."

\begin{figure}[t]

\centering

\includegraphics[width=0.92\linewidth]{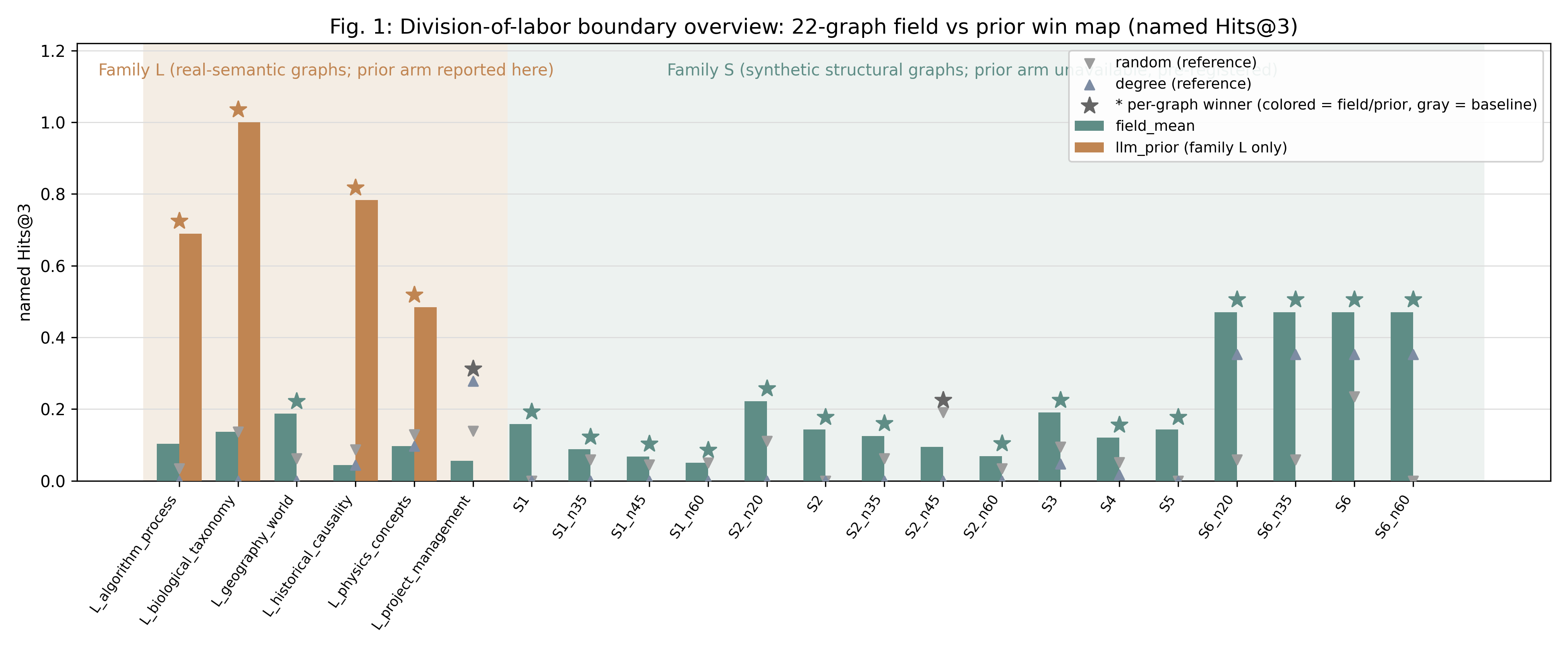}

\caption{Overview of the division-of-labor boundary (22-graph field–prior win/loss map; field-arm bars plus family-L prior-arm bars plus random/degree reference scatter; the per-graph winner is marked with ★)}

\end{figure}

\begin{figure}[t]

\centering

\includegraphics[width=0.92\linewidth]{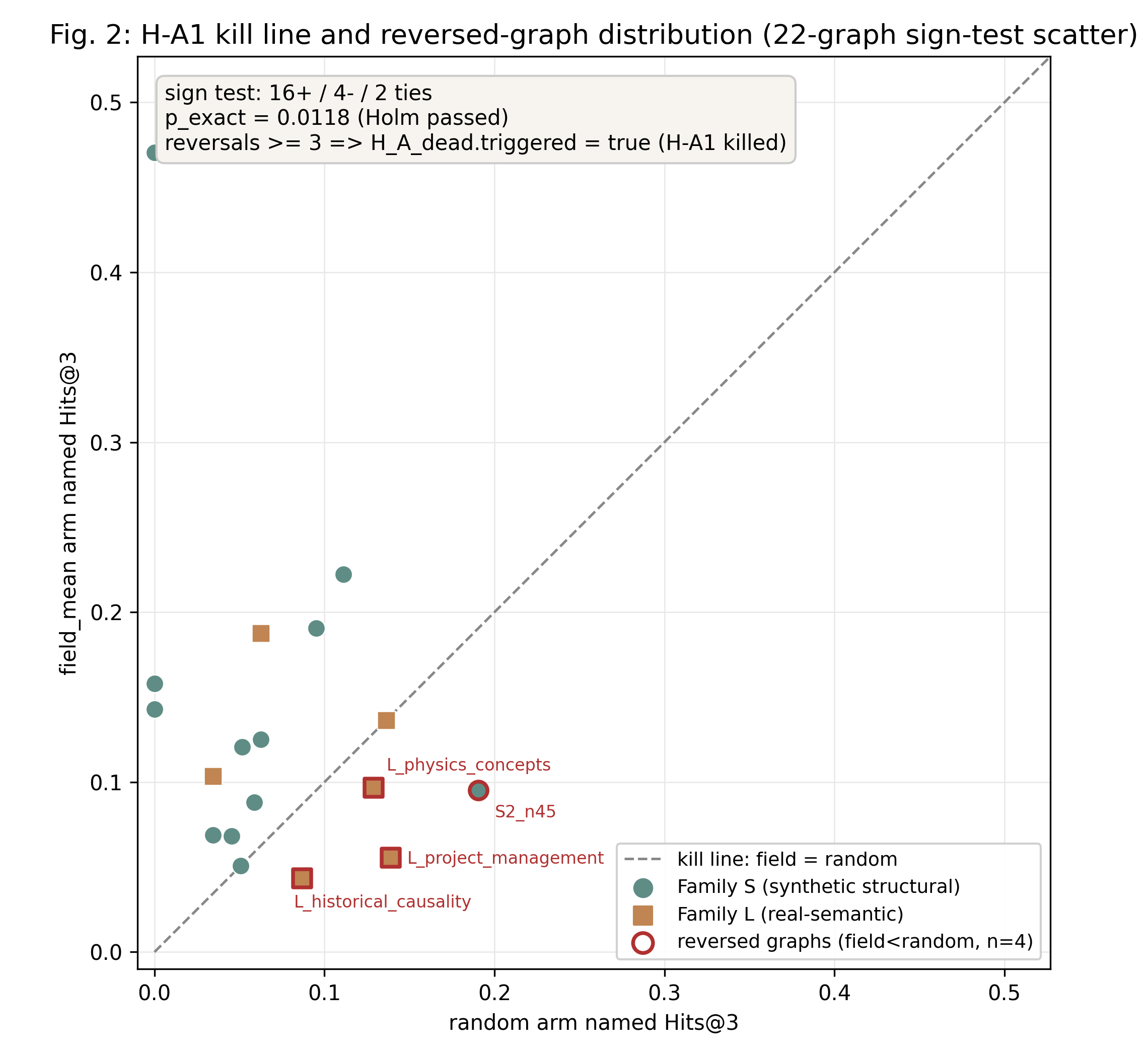}

\caption{The H-A1 kill line and the distribution of reversed graphs (22-graph sign-test scatter, n\_pos=16/n\_neg=4/n\_tie=2, the four reversed graphs marked with red boxes)}

\end{figure}

\textbf{Cross-vendor robustness (GT-3b, restricted-consistent)}: five evaluators from three model families reproduce the prior advantage on Kimi-generated graphs—doubao 4/4 and deepseek 6/6 pass the criterion, the three model families combined show 0 domains where prior ≤ field, and Kendall's W=1.0 across all ok domains (per-domain rankings in complete agreement). The hypothesis "the prior advantage is a same-vendor same-source contamination artifact" is substantially weakened; the residual limitation: all three families are Chinese-optimized large models, and shared Chinese corpora cannot be ruled out.

\textbf{Demarcation regularities}: an exploratory regression (n=20) shows hub\_concentration (max in-degree / edge count) is positively associated and the strongest correlate of field efficacy (positive direction, p=2.8×10$^{-4}$), while real\_semantics significantly suppresses field performance (negative direction, p=0.012), R$^{2}$=0.628; the coefficient point estimates are archived in Appendix A and the text keeps only direction and significance; two qualifications—n=20 is a small sample and the regression is labeled exploratory, and the features are themselves corpus-design variables, creating quasi-circularity, so the coefficients indicate directional association only. Out-of-corpus replication: on the hub axis, both paired new-graph pairs (2/2) moved in the same direction (direction only; power suffices only to resolve very large effects); on the real\_semantics axis, 3/4 domains in total satisfy "prior stronger"—GT-8b's two new domains pass the 0.6/0.2 pre-registered thresholds 2/2 (chinese\_dynasties prior 0.7805 versus field 0.0732; chemical\_elements 0.6429 versus 0.1429; an initial inconclusive was turned positive by a pre-registered amendment adding data, with the chain on record), while GT-8c's cross-backend retest is judged mixed—biological\_taxonomy prior 1.000 versus field 0.075 (diff +0.925) crosses the line, programming\_concepts prior 0.500 versus field 0.3333 (diff +0.1667) misses the margin threshold, and in GT-8c the graph-generation arm and the prior arm share the same backend and the same model, so same-source contamination risk is on record. The regularities read: high hub\_concentration → structural signal is usable overall (not exclusive to the field); real\_semantics=1 → use the semantic prior. Cross-domain heterogeneity is on record; the strength tier remains directional evidence, not extrapolated to tasks beyond family L or to larger graphs.

\begin{figure}[t]

\centering

\includegraphics[width=0.92\linewidth]{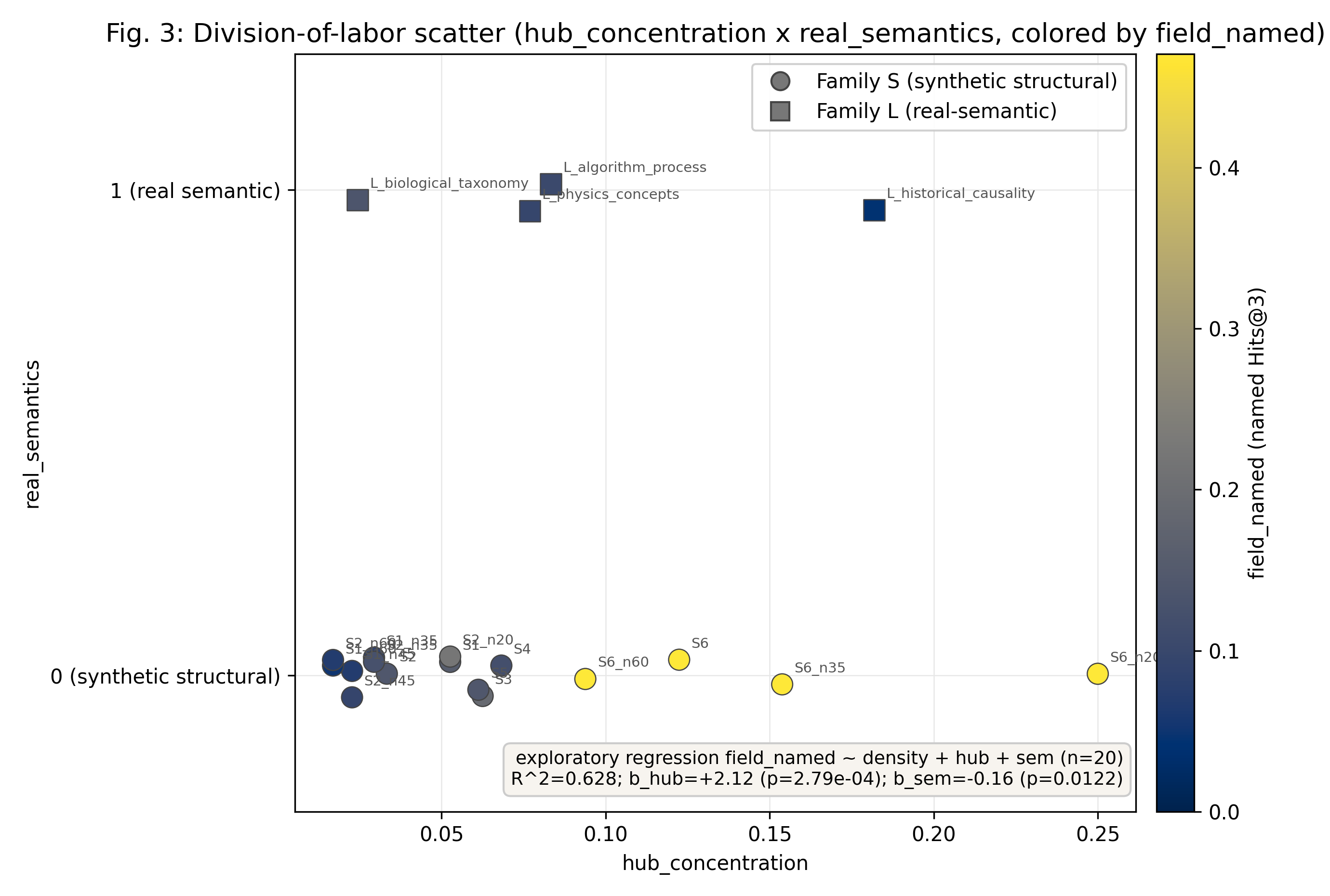}

\caption{Scatter of the division-of-labor regularities (hub\_concentration × real\_semantics, colored by field\_named; 20-graph sample; regression coefficients read from frozen fields)}

\end{figure}

\subsection{Consistency evidence and the inconclusive archive}

Beyond the main-line verdicts, four weaker pieces of evidence are archived honestly—neither upgraded nor deleted.

\textbf{GT-1 (consistency, weak discriminative power)}: Dirichlet-noise reverse is uniformly worse than the deterministic limit under the hit-rate register—dirichlet mean 0.10 versus mean-field 0.40, gap=0.30 ≥ the pre-registered line 0.2, strictly worse on 20/20 runs. A statement of discriminative power: any noise-induced degradation satisfies this pattern, so this evidence serves only as background consistency and does not independently support the potential-game reading.

\textbf{GT-2 (restricted-consistent, no\_separation)}: an attacker who knows the keyword list generates rule-evading semantic trap labels 100\% of the time (evasion\_rate=1.000, 4/4 graphs); after injecting 10 trap nodes per graph, rule\_filter drops −7.5pp on average (−10pp on three graphs, 0pp on the fourth), while field\_mean drops −0.0pp (zero on all four graphs). The pre-registered mechanical verdict is no\_separation: the rule collapse does not reach the 20pp threshold and the attack strength is not decisive, so this is not upgraded to a "rule defense fails" claim; the directional signal that the field does not read labels and is mechanistically immune to semantic traps is on record. The literature chain on the failure of rule-based defenses \cite{key28,key29,key30,key31,key32} and the methodology of adaptive attacks \cite{key33} point in the same direction, but we do not draw strength from them here.

\textbf{GT-5 and GT-2B (inconclusive archive)}: the GT-5 endpoint condition failed—the noise arm's terminal Φ exceeded mean-field on 3/4 graphs (S6 gap=−0.31), judged inconclusive and left unrewritten; we read this as "noise explores, mean-field exploits," consistent with the classical division of labor between log-linear learning \cite{key34} and deterministic best response, whereby the scattering layer's "temperature" acquires game-theoretic semantics rather than serving as a tuning knob; the narrowed monotonicity claim has since closed independently as GT-5b (§3.2). GT-2B's multi-trap strength escalation (T∈{1,2,3}) is judged inconclusive: rule\_filter scores 0.150/0.275/0.200, non-monotone, and the fixed four-option design makes the number of in-graph candidates vary inversely with T, so the field-immunity criterion is contaminated by option composition (field accuracy rises mechanically at 0.375/0.525/1.000)—the design lesson is archived as "immunity criteria must be robust to option degrees of freedom."

\textbf{GT-7 temperature frontier (mixed, the audit boundary)}: α∈{0.3,…,20} × 4 graphs × 5 seeds. The GT-5 reversal reproduces and is systematic (the same 3/4 graphs show terminal-Φ overtaking in high-temperature settings; Φ gains concentrate at the high-temperature end, corr=−0.87), so temperature genuinely controls the global exploration benefit of the potential; but a "win-win frontier" does not hold—on S6 at high temperature Φ rises while hit rate falls 0.4→0.08. Potential and hit rate are two objectives, and the audit promise covers only the former: raising temperature improves potential exploration only, not hit rate—the audit boundary is thereby drawn. Boundary-case disclosure: graphs with hit rate = 0 make the condition degenerately true; the rule was not rewritten after the fact.

\begin{figure}[t]

\centering

\includegraphics[width=0.92\linewidth]{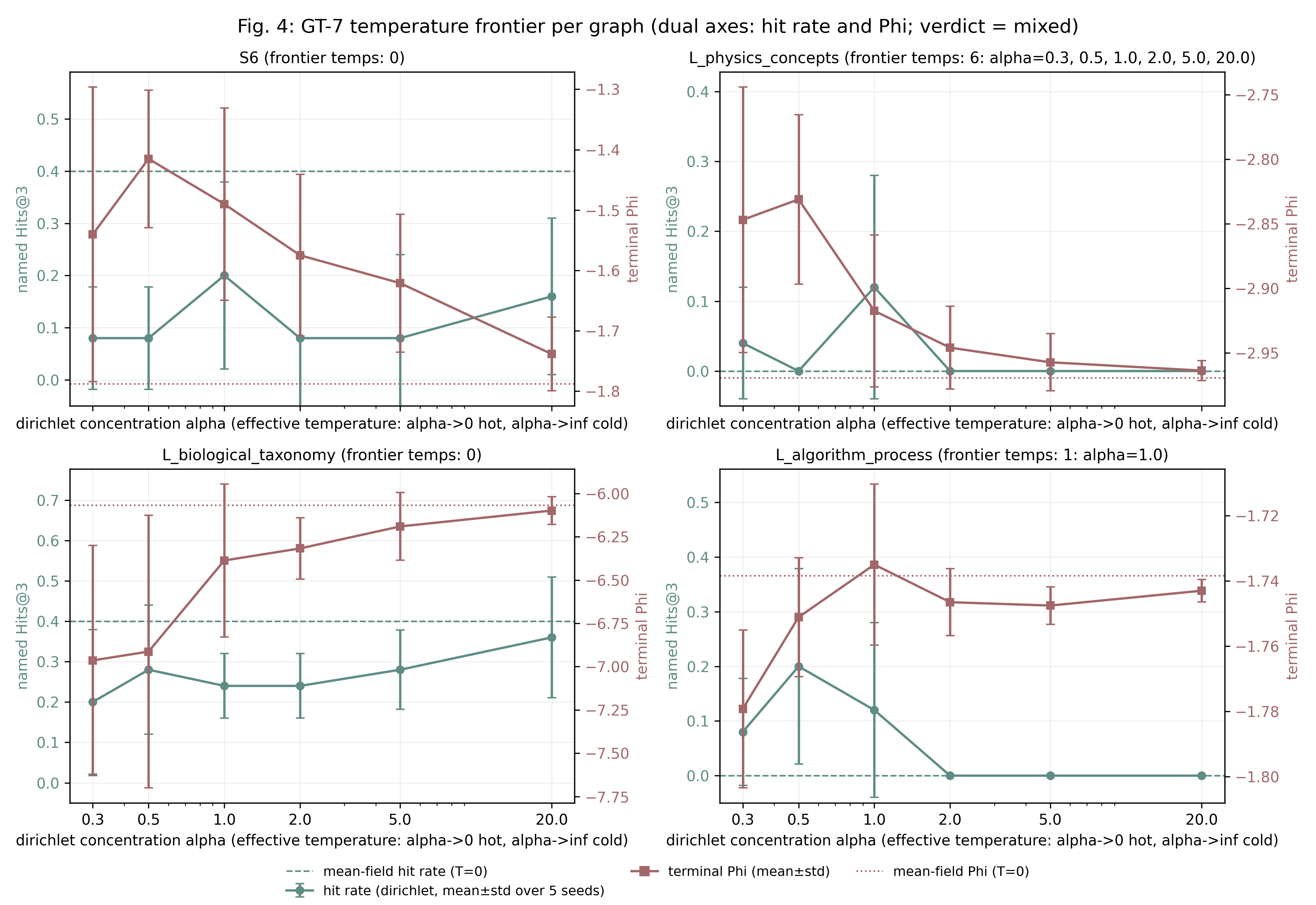}

\caption{Per-graph shape of the GT-7 temperature frontier (dual axes for hit rate and terminal Φ; 4 graphs × 6 temperature settings × 5 seeds; verdict=mixed)}

\end{figure}

\section{Honest Boundaries}

The three tiers of evidence strength are never blended in this paper. The limitations are declared centrally here, each traceable to a verdict recorded above.

\textbf{First, all three tiers of formalized equivalence are killed—the clearest limitation.} The strong formulations of "mean-field reverse = noiseless best-response dynamics" are explicitly refuted under the sampled-family/exhaustive-state kill-test (61 graphs / 338 tasks / 6760 states): P1a shows an O(1) deviation of 0.8569; P1b has min cosine −1.0 (the overshoot mechanism is on record); T-P1c is doubly killed both on the 81-point grid τ∈[0,4] and under per-state self-chosen τ* (min cos=−1.0, median τ*=0). This means none of the game-theoretic readings in this paper may rest on "the dynamics implements best response"; outside the original GT-5b/GT-6 protocols no residual claim is retained, and the positive narrative of the game-theoretic line lives only in the consistency register. The kill direction is not softened: what is refuted is not some parameter configuration but the correspondence itself.

\textbf{Second, the consistency register.} Apart from the closed pre-registered verdicts (GT-5b, GT-6) and the kill conclusions, the positive evidence is consistency evidence, not formal proof—trajectories agreeing with the potential-game reading does not amount to a proved potential-function theorem. The register is not relaxed, but neither is it upgraded to a theorem; readers should calibrate their trust in the conjunctive statements of §3.2 accordingly.

\textbf{Third, family-L graphs are generated by a single vendor's LLM.} The cross-vendor evaluator question has been addressed by GT-3b (0 defeats, W=1.0), but the single-vendor graph-generator question is not closed: the graphs' structure and label distributions carry the generating model's preferences. All three tested families are Chinese-optimized large models, same-source contamination through shared Chinese corpora cannot be ruled out, and testing non-Chinese model families and human-annotated graphs remains an open limitation.

\textbf{Fourth, the question-bank track has small samples at n=40 per cell.} ±1 problem = ±2.5pp, the same order as chance noise; every number on that track (including prior 92.5\%, field 52.5\%, rule 27.5\%) is to be read as a small-sample wide interval and is not cited as an effect size.

\textbf{Fifth, the demarcation regularities are exploratory evidence.} The n=20 regression exhibits feature–design circularity (the predictors themselves participated in corpus design); the hub axis has 2 paired pairs giving direction only, with power sufficient only for very large effects; the real\_semantics axis holds in 3/4 domains but cross-domain heterogeneity is on record (GT-8c mixed, programming\_concepts below the line). The regularities are not offered as a discriminator contribution and are not extrapolated beyond family-L tasks or to larger graphs.

\textbf{One more: the benefit of the dissipation channel remains motivation only.} GT\_FORMAL's incidental finding supplies mechanistic premise evidence for dissipation—a built-in property of the g\_a$>$0 construction (it guarantees the spectral condition ρ$<$1, avoiding divergence in 58/338 tasks)—but task-level rent remains negative on every task measured—v1\_blocking and unified tie on the synthetic benchmarks, GSM8K 86.0\%≥85.0\%, and all three arms tie at 89.9\% on StrategyQA. Mechanistic premise (guaranteeing that convergence exists, a built-in property of the construction) and task-level rent (delivering accuracy gains) are two different things: the former is closed (pre-registered); the latter is zero on every task measured in this paper, and its validation requires iterative re-search scenarios, listed as future work.

\section{Conclusion}

This paper addresses one empty layer—run-time, per-instance invariants—along two equally weighted lines of results. Auditable representation and conservation guarantees (§2): the constructive definition of three-channel scattering makes T+R+A=1 hold for arbitrary parameters, with a per-path audit residual of 2.2×10$^{-16}$ and node-by-node attributable elimination decisions; at the same time, pre-registered controls killed the attributability of the apparent accuracy advantage (E9.4/E9.5), relocating the value to machine verifiability. The game-theoretic formulation (§3): the empirical case for the auditable scalar closed (GT-5b 22/22 monotonicity, GT-6 median residual 1.594×10$^{-29}$), the distribution-level empirical coordination ratio ECR=1.333 quantifies "how far from the coordination optimum," and the temperature frontier demarcates the audit boundary; at the same time, the clearest kill conclusions sealed off all three tiers of the formalized "dynamics = best response" claim, and the P2 downgrade together with the two-directional P3 kills delineate the applicable domain of the potential explanation, incidentally yielding mechanistic premise evidence that dissipation guarantees convergence (a built-in property of the g\_a$>$0 construction). The genre discipline common to both lines: kill conclusions carry the same weight as closed (pre-registered) conclusions, the three strength tiers are never blended, and every negative result is archived without cosmetic repair. The scattering layer shows no detected accuracy difference at this sample size (E9.5: GSM8K −2pp, 95\% CI [−11.8pp, +7.8pp]; StrategyQA 0pp, 95\% CI [−8.8pp, +8.8pp], both unpaired Newcombe hybrid intervals (conservative bound; paired discordant pairs are only 0/2 and 0/0, so paired intervals degenerate; \texttt{deposon\_\allowbreak\allowbreak {}v22\_\allowbreak\allowbreak {}e95ci.json}); increments below about ±10pp cannot be excluded, and the differential-value claim rests on the E9.4 attribution mechanism); convex-combination fusion only dilutes on the measured λ settings (E9.6); "dynamics = best response" is killed by the sampled-family/exhaustive-state test—these three negatives are conclusions of this paper just as much as the three positives of the conservation ledger, potential monotonicity, and ECR quantification, not footnotes. The value proposition of the scattering layer thus compresses to one sentence: it is not more accurate, but it keeps the books, and the books can be recomputed step by step by any third party with double-precision arithmetic.

Future work, exhaustively listed: trainable KGE baselines (TransE/ComplEx/RotatE, scheduled at the 20-graph scale); re-estimating the exploratory regression after corpus expansion; upgrading the hub axis to ≥6–8 paired graph pairs for power; the question-bank track control SPEC\_GT2C and the monotone-endpoint retest SPEC\_GT5C are pre-registered and await execution. All three tiers of dynamical equivalence have been killed and are not listed as future work.

\section*{Acknowledgments}

The author thanks colleagues for discussions and feedback on early drafts.

\section{Data availability and evidence-strength conventions}

Every experimental number is traceable to a specific field of a frozen JSON under \texttt{results/}, as enumerated in Appendix~\ref{app:trace}. Every statistical verdict is the mechanical evaluation of a pre-registered decision rule (a pure function frozen before the run that reads the JSON and returns a boolean). Evidence strength is labeled in three tiers throughout: \textbf{closed (pre-registered)} (the verdict rule was frozen as a pure function before any run and mechanically evaluated to closure), \textbf{consistency evidence} (direction agrees but the pre-registered strong criterion was not met), and \textbf{motivation only} (theoretical intuition, no empirical support).

\section{Artifact availability}

All code, frozen JSONs, verdict pure functions, and the test suite ship inside the repository at \url{https://github.com/zeroandcat/Deposon}; frozen versions can be checked against the SHA-256 anchors of §3.3, and every verdict can be mechanically rerun via the bundled scripts. No proprietary dependencies are required to reproduce any number in this paper.

\section{AI use disclosure}

\label{sec:ai-use}

The Deposon framework definition and the project's research direction were provided by the corresponding author. The core algorithm, the experimental pipeline, the data-analysis scripts, and the first draft of this manuscript were produced with the assistance of KIMI-K3 (Moonshot AI) under the author's direct instruction. The author reviewed and edited every section, verified all numerical claims against the frozen JSONs in the repository, and takes full responsibility for the entire content of this paper under arXiv's authorship and AI-use policy. No content was generated by the AI without human review, and no proprietary model weights are required to reproduce any result in this paper.

\appendix

\section{Number Traceability Table}

\label{app:trace}

Every key number maps to a specific field of a frozen JSON under \texttt{results/} (verdicts are read mechanically by scripts; hand-copying is forbidden).

\begin{longtable}{p{0.35\linewidth}p{0.58\linewidth}}

\toprule

Number & Source JSON → field path \\

\midrule

\endhead

\bottomrule

\endfoot

Conservation-audit residual 2.220446049250313×10$^{-16}$, passed=\allowbreak\allowbreak {}true, tolerance 1e-6 & \texttt{deposon\_\allowbreak\allowbreak {}v19\_\allowbreak\allowbreak {}benchmark\_\allowbreak\allowbreak {}fixes.json} → \texttt{physics\_\allowbreak\allowbreak {}audit.t\_\allowbreak\allowbreak {}plus\_\allowbreak\allowbreak {}r\_\allowbreak\allowbreak {}plus\_\allowbreak\allowbreak {}a\_\allowbreak\allowbreak {}max\_\allowbreak\allowbreak {}deviation}, \texttt{physics\_\allowbreak\allowbreak {}audit.passed}, \texttt{physics\_\allowbreak\allowbreak {}audit.tolerance} \\

E9.3 post-fix GSM8K 0.82, 4 problems flipped, McNemar p=\allowbreak\allowbreak0.125; StrategyQA p=\allowbreak\allowbreak1.0 & \texttt{deposon\_\allowbreak\allowbreak {}v19\_\allowbreak\allowbreak {}benchmark\_\allowbreak\allowbreak {}fixes.json} → \texttt{experiments['E9.3\_\allowbreak\allowbreak {}high\_\allowbreak\allowbreak {}couple\_\allowbreak\allowbreak {}fix']} \\

E9.4 equal-weight control GSM8K 0.85 vs 0.04 (p=\allowbreak\allowbreak1.7e-23), StrategyQA 0.899 vs 0.202 (p=\allowbreak\allowbreak7.5e-15) & same → \texttt{experiments['E9.4\_\allowbreak\allowbreak {}equal\_\allowbreak\allowbreak {}weight\_\allowbreak\allowbreak {}decoy\_\allowbreak\allowbreak {}control'].benchmarks} \\

E9.5 rule filter GSM8K 0.87 vs 0.85 (b=\allowbreak\allowbreak0,c=\allowbreak\allowbreak2,p=\allowbreak\allowbreak0.5), StrategyQA 0.899=\allowbreak\allowbreak0.899 (p=\allowbreak\allowbreak1.0) & same → \texttt{experiments['E9.5\_\allowbreak\allowbreak {}rule\_\allowbreak\allowbreak {}baseline'].benchmarks} \\

E9.5 difference 95\% CIs: GSM8K [−11.8,+7.8]pp /\allowbreak\allowbreak StrategyQA [−8.8,+8.8]pp (unpaired Newcombe, conservative caliber); unified vs CoT (GSM8K) −12pp, CI [−20.5,−4.1]pp & \texttt{deposon\_\allowbreak\allowbreak {}v22\_\allowbreak\allowbreak {}e95ci.json} → \texttt{gsm8k}/\texttt{strategyqa}/\texttt{unified\_\allowbreak\allowbreak {}vs\_\allowbreak\allowbreak {}cot} (\texttt{method}/\texttt{ci}) \\

Synthetic benchmarks unified 100\%/\allowbreak\allowbreak100\% vs decoy-capture baseline 7\%/\allowbreak\allowbreak10\% (seed=\allowbreak\allowbreak42) & \texttt{deposon\_\allowbreak\allowbreak {}benchmark\_\allowbreak\allowbreak {}v1\_\allowbreak\allowbreak3\_\allowbreak\allowbreak {}simple.json} / \texttt{deposon\_\allowbreak\allowbreak {}benchmark\_\allowbreak\allowbreak {}v1\_\allowbreak\allowbreak3\_\allowbreak\allowbreak {}traps.json} → \texttt{variant\_\allowbreak\allowbreak {}results} \\

Label-permutation ablation 17.2\%±6.4\%; uniform-parameter degeneration 10\% (trap set) & \texttt{deposon\_\allowbreak\allowbreak {}benchmark\_\allowbreak\allowbreak {}v1\_\allowbreak\allowbreak3\_\allowbreak\allowbreak {}labelshuffle.json} → \texttt{label\_\allowbreak\allowbreak {}shuffle} / \texttt{uniform\_\allowbreak\allowbreak {}params} \\

GSM8K: CoT 97.0\%, unified 85.0\%, v1\_blocking 86.0\%, p=\allowbreak\allowbreak4.9e-4 & \texttt{deposon\_\allowbreak\allowbreak {}benchmark\_\allowbreak\allowbreak {}v1\_\allowbreak\allowbreak4\_\allowbreak\allowbreak {}gsm8k.json} \\

StrategyQA: unified 89.9\% vs CoT 92.9\% (p=\allowbreak\allowbreak0.549), three arms tied at 89.9\% & \texttt{deposon\_\allowbreak\allowbreak {}benchmark\_\allowbreak\allowbreak {}v1\_\allowbreak\allowbreak4\_\allowbreak\allowbreak {}strategyqa.json} \\

Fusion dilution: physics 0.484→\allowbreak\allowbreak0.452, historical 0.783→\allowbreak\allowbreak0.739 (λ=\allowbreak\allowbreak0.5 convex combination) & \texttt{deposon\_\allowbreak\allowbreak {}v20\_\allowbreak\allowbreak {}crossval.json} → per-graph fields of \texttt{hybrid\_\allowbreak\allowbreak {}lambda\_\allowbreak\allowbreak {}convex=0.5} \\

Four-λ single-graph scan: all four settings identical, named Hits@3=\allowbreak\allowbreak0.294, any\_lambda\_pass=\allowbreak\allowbreak {}false & \texttt{deposon\_\allowbreak\allowbreak {}v16\_\allowbreak\allowbreak {}llm\_\allowbreak\allowbreak {}prior.json} → \texttt{success\_\allowbreak\allowbreak {}evaluation} \\

λ=\allowbreak\allowbreak2 normalized-variant apparent 0.471; E9.6 null ablation 0.1176=\allowbreak\allowbreak0.1176, random edges 0 on all 5 runs & \texttt{deposon\_\allowbreak\allowbreak {}v17\_\allowbreak\allowbreak {}fusion\_\allowbreak\allowbreak {}fix.json} → \texttt{hybrid\_\allowbreak\allowbreak {}norm@2.0}; \texttt{deposon\_\allowbreak\allowbreak {}v19\_\allowbreak\allowbreak {}quickwins.json} → \texttt{E9\_\allowbreak\allowbreak6c\_\allowbreak\allowbreak {}lambda2\_\allowbreak\allowbreak {}null\_\allowbreak\allowbreak {}ablation} \\

H-A1 killed: 16+/\allowbreak\allowbreak4−/\allowbreak\allowbreak2, p=\allowbreak\allowbreak0.0118, kill line triggered, 4 reversed graphs & \texttt{deposon\_\allowbreak\allowbreak {}v20\_\allowbreak\allowbreak {}corpus\_\allowbreak\allowbreak {}eval.json} → \texttt{verdicts.H\_\allowbreak\allowbreak {}A1\_\allowbreak\allowbreak {}field\_\allowbreak\allowbreak {}mean\_\allowbreak\allowbreak {}gt\_\allowbreak\allowbreak {}random.sign\_\allowbreak\allowbreak {}test}, \texttt{verdicts.kill\_\allowbreak\allowbreak {}lines.H\_\allowbreak\allowbreak {}A\_\allowbreak\allowbreak {}dead} \\

H-A2 survives: 19+/\allowbreak\allowbreak1−/\allowbreak\allowbreak2, p=\allowbreak\allowbreak4.0e-5 & same → \texttt{verdicts.H\_\allowbreak\allowbreak {}A2\_\allowbreak\allowbreak {}field\_\allowbreak\allowbreak {}mean\_\allowbreak\allowbreak {}gt\_\allowbreak\allowbreak {}degree.sign\_\allowbreak\allowbreak {}test.p\_\allowbreak\allowbreak {}exact=4.005e-05} \\

Wilcoxon p=\allowbreak\allowbreak0.0031/\allowbreak\allowbreak $|r|$ =\allowbreak\allowbreak0.83; paired t p$<$0.0001/\allowbreak\allowbreak {}d=\allowbreak\allowbreak2.05 & \texttt{v20\_\allowbreak\allowbreak {}statcheck\_\allowbreak\allowbreak {}fm\_\allowbreak\allowbreak {}vs\_\allowbreak\allowbreak {}rand.json} (p\_value=0.003052), \texttt{v20\_\allowbreak\allowbreak {}statcheck\_\allowbreak\allowbreak {}fm\_\allowbreak\allowbreak {}vs\_\allowbreak\allowbreak {}deg.json} (p=2.13e-08, d=2.0478) \\

Regression β(hub)=\allowbreak\allowbreak2.12 (p=\allowbreak\allowbreak2.8e-4), β(real\_sem)=\allowbreak\allowbreak−0.16 (p=\allowbreak\allowbreak0.012), R$^{2}$=\allowbreak\allowbreak0.628, n=\allowbreak\allowbreak20 & \texttt{v20\_\allowbreak\allowbreak {}regression\_\allowbreak\allowbreak {}field\_\allowbreak\allowbreak {}v2.json} → \texttt{coefficients.*}, \texttt{r\_\allowbreak\allowbreak {}squared=0.628226}, \texttt{n\_\allowbreak\allowbreak {}observations=20} \\

GT-1 gap=\allowbreak\allowbreak0.30 (0.10 vs 0.40), strictly worse on 20/\allowbreak\allowbreak20 runs & \texttt{deposon\_\allowbreak\allowbreak {}v20\_\allowbreak\allowbreak {}gt.json} → \texttt{GT1\_\allowbreak\allowbreak {}potential\_\allowbreak\allowbreak {}game\_\allowbreak\allowbreak {}convergence.verdict} \\

GT-4 ECR median 1.333 (17 graphs) /\allowbreak\allowbreak 1.5 (family S, 13 graphs); family L 0.5/\allowbreak\allowbreak0.75; ∞×3 counted separately, excluded from the median; S2\_n45=\allowbreak\allowbreak0.5 (field names \texttt{GT4\_\allowbreak\allowbreak {}price\_\allowbreak\allowbreak {}of\_\allowbreak\allowbreak {}anarchy} and \texttt{field\_\allowbreak\allowbreak {}coordination\_\allowbreak\allowbreak {}value\_\allowbreak\allowbreak {}supported} are frozen conventions, not renamed; coverage 20/\allowbreak\allowbreak22 graphs: L\_geography\_world and L\_project\_management not evaluated in the frozen run—arm data absent, neither finite nor ∞) & same → \texttt{GT4\_\allowbreak\allowbreak {}price\_\allowbreak\allowbreak {}of\_\allowbreak\allowbreak {}anarchy.verdict.poa\_\allowbreak\allowbreak {}per\_\allowbreak\allowbreak {}graph\_\allowbreak\allowbreak {}finite}, \texttt{n\_\allowbreak\allowbreak {}poa\_\allowbreak\allowbreak {}inf=3} \\

GT-5b 22/\allowbreak\allowbreak22 monotone (monotonicity rate 100\%, pre-registered line 80\%) & \texttt{deposon\_\allowbreak\allowbreak {}v20\_\allowbreak\allowbreak {}gt5b.json} → \texttt{per\_\allowbreak\allowbreak {}graph\_\allowbreak\allowbreak {}summary.*.meanfield\_\allowbreak\allowbreak {}monotone\_\allowbreak\allowbreak {}rate=1.0} \\

GT-5 endpoint reversal S6 gap=\allowbreak\allowbreak−0.31 (3/\allowbreak\allowbreak4 graphs inconclusive) & \texttt{deposon\_\allowbreak\allowbreak {}v20\_\allowbreak\allowbreak {}gt5.json} → \texttt{per\_\allowbreak\allowbreak {}graph\_\allowbreak\allowbreak {}detail.S6} \\

GT-6 median residual 1.594e-29; exceptions S4=\allowbreak\allowbreak0.148 /\allowbreak\allowbreak L\_algorithm\_process=\allowbreak\allowbreak0.136 /\allowbreak\allowbreak S5=\allowbreak\allowbreak0.121 & \texttt{deposon\_\allowbreak\allowbreak {}v20\_\allowbreak\allowbreak {}gt6.json} → \texttt{verdict.median\_\allowbreak\allowbreak {}residual\_\allowbreak\allowbreak {}ratio}, \texttt{per\_\allowbreak\allowbreak {}graph\_\allowbreak\allowbreak {}summary.*.residual\_\allowbreak\allowbreak {}ratio\_\allowbreak\allowbreak {}mean} \\

GT-7 mixed: corr −0.87; S6 hit rate 0.4→\allowbreak\allowbreak0.08 & \texttt{deposon\_\allowbreak\allowbreak {}v20\_\allowbreak\allowbreak {}gt7.json} → \texttt{per\_\allowbreak\allowbreak {}graph} (corr −0.87 is the document-level summary value (GT\_RECONSTRUCTION §7), not a direct field read) \\

GT-2 no\_separation: rule −7.5pp (0.1/\allowbreak\allowbreak0.0/\allowbreak\allowbreak0.1/\allowbreak\allowbreak0.1), field −0.0pp, evasion=\allowbreak\allowbreak1.0 & \texttt{deposon\_\allowbreak\allowbreak {}v20\_\allowbreak\allowbreak {}crossval.json} → \texttt{gt2\_\allowbreak\allowbreak {}verdict}, \texttt{gt2\_\allowbreak\allowbreak {}attacker\_\allowbreak\allowbreak {}meta.*.evasion\_\allowbreak\allowbreak {}rate=1.0} \\

GT-2B inconclusive: rule 0.150/\allowbreak\allowbreak0.275/\allowbreak\allowbreak0.200; field 0.375/\allowbreak\allowbreak0.525/\allowbreak\allowbreak1.000 & \texttt{deposon\_\allowbreak\allowbreak {}v20\_\allowbreak\allowbreak {}gt2b.json} → \texttt{verdict}, \texttt{per\_\allowbreak\allowbreak {}T.*.per\_\allowbreak\allowbreak {}domain.*.accuracy} \\

GT-3b: doubao 4/\allowbreak\allowbreak4, deepseek 6/\allowbreak\allowbreak6, 0 defeats, Kendall W=\allowbreak\allowbreak1.0 & \texttt{deposon\_\allowbreak\allowbreak {}v20\_\allowbreak\allowbreak {}gt3.json} → \texttt{verdict.H\_\allowbreak\allowbreak {}GT3\_\allowbreak\allowbreak {}supported=true}, top-level \texttt{kendall\_\allowbreak\allowbreak {}W=1.0} \\

GT-8 2/\allowbreak\allowbreak2 same direction (pair A +0.7917$>$+0.1333; pair B +0.0526$>$−0.0833) & \texttt{deposon\_\allowbreak\allowbreak {}v20\_\allowbreak\allowbreak {}gt8.json} → \texttt{verdict.verdict="supports\_\allowbreak\allowbreak {}H\_\allowbreak\allowbreak {}GT8"}, \texttt{per\_\allowbreak\allowbreak {}pair.*} \\

GT-8b supports\_H\_GT8B: chinese\_dynasties 0.7805 vs 0.0732 (diff +0.7073); chemical\_elements 0.6429 vs 0.1429 (diff +0.5000) & \texttt{deposon\_\allowbreak\allowbreak {}v20\_\allowbreak\allowbreak {}gt8b.json} → \texttt{gt8b\_\allowbreak\allowbreak {}verdict}, \texttt{per\_\allowbreak\allowbreak {}domain.*.named\_\allowbreak\allowbreak {}summary}, \texttt{prior\_\allowbreak\allowbreak {}named\_\allowbreak\allowbreak {}minus\_\allowbreak\allowbreak {}field\_\allowbreak\allowbreak {}named} \\

GT-8c mixed: biological\_taxonomy 1.000 vs 0.075 (diff +0.925) crosses; programming\_concepts 0.500 vs 0.3333 (diff +0.1667) misses; vendor=\allowbreak\allowbreak {}volces\_ark\_bytedance & \texttt{deposon\_\allowbreak\allowbreak {}v20\_\allowbreak\allowbreak {}gt8c.json} → \texttt{gt8c\_\allowbreak\allowbreak {}verdict}, \texttt{per\_\allowbreak\allowbreak {}domain.*.named\_\allowbreak\allowbreak {}summary}, \texttt{backend.*} \\

GT\_FORMAL: P1a max‖T−BR‖∞=\allowbreak\allowbreak0.8569, 1−cos≈0.91 does not vanish with lr; P1b min cos=\allowbreak\allowbreak−1.0, min ΔΦ=\allowbreak\allowbreak−1.2424e−2; P2 acyclic r≤5.9e−16, cyclic median 0.669, fraction 0.924; P3 max r difference 0.1585, fixed-point difference 0.8442, g\_a=\allowbreak\allowbreak0 divergence in 58 tasks; 61 graphs/\allowbreak\allowbreak338 tasks/\allowbreak\allowbreak6760 states & \texttt{deposon\_\allowbreak\allowbreak {}v21\_\allowbreak\allowbreak {}gtformal.json} (seed=210021) → \texttt{verdict.T\_\allowbreak\allowbreak {}P1b}, \texttt{verdict.P1a\_\allowbreak\allowbreak {}deviation}, \texttt{verdict.T\_\allowbreak\allowbreak {}P2}, \texttt{verdict.T\_\allowbreak\allowbreak {}P3}, \texttt{residuals.dag}, \texttt{residuals.cyclic}, \texttt{n\_\allowbreak\allowbreak {}graphs/n\_\allowbreak\allowbreak {}tasks/n\_\allowbreak\allowbreak {}states} \\

T-P1c killed: 81-point grid over τ∈[0,4] fails everywhere, min cos=\allowbreak\allowbreak−1.0 (both strong form and per-state τ*), median τ*=\allowbreak\allowbreak0 (frac\_at\_tau0=\allowbreak\allowbreak0.8598) & \texttt{deposon\_\allowbreak\allowbreak {}v22\_\allowbreak\allowbreak {}p1c.json} (seed=210021) → \texttt{verdict}, \texttt{descriptive.tau\_\allowbreak\allowbreak {}star\_\allowbreak\allowbreak {}per\_\allowbreak\allowbreak {}state}, \texttt{tau\_\allowbreak\allowbreak {}grid} \\

Three limiting-state per-path average dissipation 0 /\allowbreak\allowbreak 3.63 /\allowbreak\allowbreak 0.358 (energy units) & \texttt{deposon\_\allowbreak\allowbreak {}benchmark\_\allowbreak\allowbreak {}v1\_\allowbreak\allowbreak3\_\allowbreak\allowbreak {}traps.json} → \texttt{variant\_\allowbreak\allowbreak {}results.{v1\_\allowbreak\allowbreak {}blocking,v2\_\allowbreak\allowbreak {}tunneling,unified}.avg\_\allowbreak\allowbreak {}ether\_\allowbreak\allowbreak {}dissipated} \\

\end{longtable}

\section{Run-Level Verdict Ledger of the Deposon Program (V3--V4)}

\label{app:ledger}


\subsection*{B.0 Scope and Positioning}
This appendix is a run-level evidence ledger supporting the mainline's auditability claims: it registers, in ledger form, the verdicts, kill dispositions, and re-verification chains of the program's V3 and V4 follow-on families of the mainline's Sections 2--3, as a continuation of the program's run-level record. Here ``V3'' and ``V4'' name the program's two follow-on families---the V3 side is the game-theoretic adjudication line registered in the record's Section 4.7, and the V4 side is the critical-learning line of the record's Sections 4.1--4.6; the numbering is the program's own run-level naming, carried as such. The ledger's rows carry tiered depth, in the record's own tiering: the T1 verdict line (verdict wording plus key frozen numbers, carried literally), the T2 numeric-detail paragraph (frozen values verbatim), the T3 carrier anchor (path plus SHA-256 anchor and byte count), and the T4 register-level single line; tiers mark depth of registration, not weight of evidence. Carrier anchors are SHA-256 anchors (first 12 hex digits), in the mainline's Section 3.3 wording. Time basis: every measurement, verdict, and anchor in this appendix is stated as of its own registration date (2026-09-16 through 2026-10-29); no claim of currency beyond those dates is made.

Forward-looking note (direction level): a follow-on line of the program is at an early pre-research stage on binding the ledger-layer invariants of this discipline to operations at the parameter/weight layer of the underlying models---the weight-space programming direction; no proven dependency, result, or instrument from that line is claimed in this revision, and the note is registered here as direction only.

\subsection*{B.1 Availability Grading Notes}
Anchor paths in this appendix are pointers into the program's registered artifact faces, not publication claims; paths carrying the (project-internal) qualifier are project-internal working paths and are not asserted to exist on any public face. The ledger's carrier column carries each row's availability grade: rows marked \emph{repo} anchor files on the public repository face; rows marked \emph{sub} anchor files in the program's sub-archives (project-internal working faces); rows at \emph{register level} anchor a group-level carrier whose per-file fingerprints are derivable from the register index (the errata chain and the inventory govern); [L]-marked rows are held on the author's local face, outside the public workspace, per the register's own note. This grading is the successor statement to the mainline's artifact-availability wording: the anchored artifacts remain at their registered availability faces, no new publication is claimed by this revision, and the appendix's carrier claims are grade-scoped, not universal-availability claims.

\subsection*{B.2 V3 Family Ledger (24 Lines)}
\begin{longtable}{p{0.06\linewidth}p{0.33\linewidth}p{0.10\linewidth}p{0.39\linewidth}}
\toprule
Ledger line & Line identifier & Tier & Frozen-value anchors \\
\midrule
\endhead
\bottomrule
\endfoot
1 & V3-01 30-cell four-value identifiability & T1+\allowbreak\allowbreak {}T2+\allowbreak\allowbreak {}T3 & \textbf{kd} (known defects; PI-authorized 2026-10-28); max\_u=1 fixed; L15+U15 underivable; carrier \texttt{\_\allowbreak\allowbreak {}v3\_\allowbreak\allowbreak {}01\_\allowbreak\allowbreak {}verdict\_\allowbreak\allowbreak {}2026\_\allowbreak\allowbreak {}10\_\allowbreak\allowbreak {}28.md} (85,820 B; SHA-12 4cb27e735283; repo). \\
2 & V3-02 ETHard closure (U-V04-01) & T1+T3 & Output layer half closed; 11/11 K-PV02 PASS; protocol\_version V0.4; U-V04-02 unopened; carrier \texttt{\_\allowbreak\allowbreak {}v3\_\allowbreak\allowbreak {}02\_\allowbreak\allowbreak {}ethard\_\allowbreak\allowbreak {}closure\_\allowbreak\allowbreak {}verdict\_\allowbreak\allowbreak {}2026\_\allowbreak\allowbreak {}10\_\allowbreak\allowbreak {}09.md} (27,819 B; SHA-12 aa0f96bb618f; sub; R27-corrected). \\
3 & V3 supplement (P-A five leads) & T1+\allowbreak\allowbreak {}T2+\allowbreak\allowbreak {}T3 & C1 degenerate (established); C2 non-degenerate; C3/C4/C5 degenerate; sim\_rm=6; rbr\{2,200\}; bayes\{1,200\}; LSH 22 $\to$ 6 codes, 62/231 pairs Hamming=0; carrier \texttt{\_\allowbreak\allowbreak {}v3\_\allowbreak\allowbreak {}supplement\_\allowbreak\allowbreak {}verdict\_\allowbreak\allowbreak {}2026\_\allowbreak\allowbreak {}09\_\allowbreak\allowbreak {}23.md} (15,247 B; SHA-12 b70211bfa83e; sub). \\
4 & V3-R re-check register (13 lines) & T1+T3 & Two-pass re-adjudication; all seven sign-offs closed (R-1 through R-7, the re-adjudication chain locked; cross-family closures covering the register's rows, not a seventh-per-line count); 13 lines in one table; 7/7 signed; carrier \texttt{\_\allowbreak\allowbreak {}v3\_\allowbreak\allowbreak {}recheck\_\allowbreak\allowbreak {}verdict\_\allowbreak\allowbreak {}register\_\allowbreak\allowbreak {}2026\_\allowbreak\allowbreak {}09\_\allowbreak\allowbreak {}27.md} (112,085 B; SHA-12 d9b6b974cfc1; repo). \\
5 & V3 re-check 26b & T1+T3 & Temporal requirement established (semantic layer separately judged); wording 2h18m ahead of all measurements; 2d5h35m ahead of standing readings; carrier \texttt{\_\allowbreak\allowbreak {}v3\_\allowbreak\allowbreak {}recheck\_\allowbreak\allowbreak {}26b\_\allowbreak\allowbreak {}verdict\_\allowbreak\allowbreak {}2026\_\allowbreak\allowbreak {}10\_\allowbreak\allowbreak {}01.md} (26,309 B; SHA-12 8bbecb15b28b; sub). \\
6 & V3-N three indeterminates (\#19/\#21/\#32) & T1+T3 & Resolved: \#19 run-established (P-G boss\_pg, PASS 2/GRAY 1/FAIL 0, the indeterminate lifted); \#21 root-cause substituted (KT-B1); \#32 externally blocked; no new PI-rewritable evidence from any of the three; carrier \texttt{\_\allowbreak\allowbreak {}v3\_\allowbreak\allowbreak {}n\_\allowbreak\allowbreak {}recheck\_\allowbreak\allowbreak {}llm\_\allowbreak\allowbreak {}verdict\_\allowbreak\allowbreak {}2026\_\allowbreak\allowbreak {}09\_\allowbreak\allowbreak {}27.md} (21,332 B; SHA-12 426cfe18cf65; repo). \\
7 & V3N cpath live re-run (\#32) & T1+\allowbreak\allowbreak {}T2+\allowbreak\allowbreak {}T3 & Judgment band 0 new settings (per de772cd9e7ba \S5); C-path RAG theoretical margin \textbf{0\% to +26.7\%}; runner 4 changes; 20-line unified diff; glm-latest plus doubao-embedding-vision-251215; carrier \texttt{\_\allowbreak\allowbreak {}v3n\_\allowbreak\allowbreak {}cpath\_\allowbreak\allowbreak {}live\_\allowbreak\allowbreak {}rerun\_\allowbreak\allowbreak {}verdict\_\allowbreak\allowbreak {}2026\_\allowbreak\allowbreak {}10\_\allowbreak\allowbreak {}01.md} (16,409 B; SHA-12 a91efbfec3d0; sub; data 7c26d08ece3b). \\
8 & K-V3DF-0-1 readings closure & T1+T3 & K-V3R-0-A \textbf{PASS} (anti-degeneration gate: n\_distinct$>$3 and std$>$0, four readings complete, zero claims of full-chain coverage); carrier \texttt{\_\allowbreak\allowbreak {}k\_\allowbreak\allowbreak {}v3df\_\allowbreak\allowbreak {}0\_\allowbreak\allowbreak {}1\_\allowbreak\allowbreak {}readings\_\allowbreak\allowbreak {}verdict\_\allowbreak\allowbreak {}2026\_\allowbreak\allowbreak {}10\_\allowbreak\allowbreak {}09.md} (37,003 B; SHA-12 e98bee353448; sub; R27-corrected, original face). \\
9 & K-V3DF-9-1 C1--C4 legs & T1+T3 & Per the verdict document (authoritative source): \textbf{KD} (not adjudicable; C1--C4 legs, faces 1--2); K-V3DF-4-2 sustains the prior FAIL; zero re-grading; carrier \texttt{\_\allowbreak\allowbreak {}k\_\allowbreak\allowbreak {}v3df\_\allowbreak\allowbreak {}9\_\allowbreak\allowbreak {}1\_\allowbreak\allowbreak {}c1\_\allowbreak\allowbreak {}c4\_\allowbreak\allowbreak {}legs\_\allowbreak\allowbreak {}verdict\_\allowbreak\allowbreak {}2026\_\allowbreak\allowbreak {}10\_\allowbreak\allowbreak {}09.md} (27,735 B; SHA-12 a14aafa220c4; sub). \\
\multicolumn{4}{p{0.92\linewidth}}{\footnotesize\emph{Erratum (row 9, registered explicitly): the record's inventory carries a cross-row mis-attribution -- the D4 dual-factor re-label and the two judgment faces 735c03db1ae9 (v3r1) / 585714f9660c (v3), listed under this row in the record's Ledger row 9 and \S4.7 line 412, belong to the Task B v3 chain (row 14 of this table). Correction basis: the inquiry receipt, section 3 (8cf623a06a67); verdict document measured 27,735 B, SHA-12 A14AAFA220C4, byte-identical via public-mirror \texttt{git show}; in-document hits for D4 / pi\_cot / 585714 / 735c03 / re-label: zero.}} \\
10 & K-V3S-4-2 domain extension & T1+T3 & Registered; carrier \texttt{\_\allowbreak\allowbreak {}k\_\allowbreak\allowbreak {}v3s\_\allowbreak\allowbreak {}4\_\allowbreak\allowbreak {}2\_\allowbreak\allowbreak {}domain\_\allowbreak\allowbreak {}extension\_\allowbreak\allowbreak {}judgment\_\allowbreak\allowbreak {}2026\_\allowbreak\allowbreak {}10\_\allowbreak\allowbreak {}09.md} (23,018 B; SHA-12 a802af7ac3bc; sub). \\
11 & E6 TH V3 (19 residual exits) & T1+T3 & Rationale-chain defect, not four-value; 19 exits; $\eta$\_v3 readings on record; carrier \texttt{\_\allowbreak\allowbreak {}e6\_\allowbreak\allowbreak {}th\_\allowbreak\allowbreak {}v3\_\allowbreak\allowbreak {}19\_\allowbreak\allowbreak {}residual\_\allowbreak\allowbreak {}exits\_\allowbreak\allowbreak {}judgment\_\allowbreak\allowbreak {}2026\_\allowbreak\allowbreak {}10\_\allowbreak\allowbreak {}10.md} (46,261 B; SHA-12 442d5ff4acdb; sub). \\
12 & E6 V3 rationale review & T1+T3 & Rationale gap confirmed; carrier \texttt{\_\allowbreak\allowbreak {}e6\_\allowbreak\allowbreak {}verdict\_\allowbreak\allowbreak {}v3\_\allowbreak\allowbreak {}rationale\_\allowbreak\allowbreak {}review\_\allowbreak\allowbreak {}2026\_\allowbreak\allowbreak {}10\_\allowbreak\allowbreak {}10.md} (39,008 B; SHA-12 664e82968fc2; sub). \\
13 & coze C-5 evidence verification & T1+T3 & Four items FOUND\_IN\_REPO (zero ghost citations); naming-face anomaly registered as-is (the v1 entity files carry the \texttt{\_v4\_} prefix); carrier \texttt{\_\allowbreak\allowbreak {}coze\_\allowbreak\allowbreak {}v1\_\allowbreak\allowbreak {}v3\_\allowbreak\allowbreak {}core\_\allowbreak\allowbreak {}principle\_\allowbreak\allowbreak {}verification\_\allowbreak\allowbreak {}2026\_\allowbreak\allowbreak {}10\_\allowbreak\allowbreak {}18.md} (30,800 B; SHA-12 dfcda086b498; repo). \\
14 & Task B v3 pre-registration (pi\_cot v3) & T1+T3 & Registered, unadjudicated in this phase; D4 dual-factor re-labels; lock-and-rejudge and missing-ledger corrections; result\_v3 585714f9660c / v3r1 735c03db1ae9 (two judgment faces); [the unique double-counted entry of the coverage accounting: 37 file-level entries $\to$ 36 lines (B.6)]; carrier \texttt{\_\allowbreak\allowbreak {}v4\_\allowbreak\allowbreak {}pi\_\allowbreak\allowbreak {}cot\_\allowbreak\allowbreak {}v3\_\allowbreak\allowbreak {}verdict\_\allowbreak\allowbreak {}v3.md} (53,751 B; SHA-12 bb44fdc7ab0f; repo; L1/V3 archive; one pre-registration generation beyond the v2.2 instance of \S4.2). \\
15 & V5 GT series (GT5/GT6/GT8/GT5b/GT) & T1+T3 & Formal closure (three-state frozen protocol K-V5R3-0-A: PASS/FAIL/KD only, no prose verdicts); executor 9a679f43f796 + 8 execution artifacts; carrier \texttt{\_\allowbreak\allowbreak {}v5\_\allowbreak\allowbreak {}gt\_\allowbreak\allowbreak {}verdict\_\allowbreak\allowbreak {}2026\_\allowbreak\allowbreak {}09\_\allowbreak\allowbreak {}29.md} (57,227 B; SHA-12 45648b0ec940; repo); [cross-family: L1/V3 archive storage, V5 carrier naming, V3-side experimental semantics; disclosed in the coverage accounting, not double-counted]. \\
16 & GT-8 P-10 low arm (r2 best-gate) & T1+T3 & Terminal update; independent measurement face; the gate-passing set governing; carrier \texttt{\_\allowbreak\allowbreak {}v5\_\allowbreak\allowbreak {}gt8\_\allowbreak\allowbreak {}lowarm\_\allowbreak\allowbreak {}verdict\_\allowbreak\allowbreak {}r2\_\allowbreak\allowbreak {}2026\_\allowbreak\allowbreak {}09\_\allowbreak\allowbreak {}29.md} (40,282 B; SHA-12 aff727434aa4; sub); [cross-family annotation as row 15]. \\
17 & item35 v2 matrix & T1+T3 & Registered; carrier \texttt{\_\allowbreak\allowbreak {}v5\_\allowbreak\allowbreak {}item35\_\allowbreak\allowbreak {}v2\_\allowbreak\allowbreak {}verdict\_\allowbreak\allowbreak {}2026\_\allowbreak\allowbreak {}09\_\allowbreak\allowbreak {}29.md} (26,232 B; SHA-12 b59bef293712; repo); [cross-family annotation as row 15]. \\
18 & item35 v3 collinear & T1+T3 & Registered (2026-10-01 recompute; its recompute JSON on record); carrier \texttt{\_\allowbreak\allowbreak {}v5\_\allowbreak\allowbreak {}item35\_\allowbreak\allowbreak {}v3\_\allowbreak\allowbreak {}collinear\_\allowbreak\allowbreak {}verdict\_\allowbreak\allowbreak {}2026\_\allowbreak\allowbreak {}10\_\allowbreak\allowbreak {}01.md} (38,266 B; SHA-12 363cce3dae59; sub; R27-corrected -- the record's printed digest 873652fd39dd/42,504 belongs to the prereg file); [cross-family annotation as row 15]. \\
19 & pjpm\_pc bang3 & T1+\allowbreak\allowbreak {}T2+\allowbreak\allowbreak {}T3 & Total verdict 2: \textbf{FAIL}; K-PJPC-0-1 with no missing items, coverage true, FAIL standing; three prior archives with zero overrides, zero withdrawals, zero retroactive admissions; carrier \texttt{\_\allowbreak\allowbreak {}v5\_\allowbreak\allowbreak {}pjpm\_\allowbreak\allowbreak {}pc\_\allowbreak\allowbreak {}verdict\_\allowbreak\allowbreak {}bang3\_\allowbreak\allowbreak {}2026\_\allowbreak\allowbreak {}09\_\allowbreak\allowbreak {}29.md} (41,566 B; SHA-12 1814f9590f0f; sub); [cross-family annotation as row 15]. \\
20 & V3 deg P-1 re-open (r2) & T1+T3 & Three-state terminal; $\gamma$ root-cause column; T1 adjudicated cause; name collision handled by issuing a new name, zero overwrites; kill-line 98286cc1aec7; execution 38231ef6f053; re-verify 4afeb8cd41de; carrier \texttt{\_\allowbreak\allowbreak {}v5\_\allowbreak\allowbreak {}v3\_\allowbreak\allowbreak {}deg\_\allowbreak\allowbreak {}p1\_\allowbreak\allowbreak {}reopen\_\allowbreak\allowbreak {}verdict\_\allowbreak\allowbreak {}r2\_\allowbreak\allowbreak {}2026\_\allowbreak\allowbreak {}09\_\allowbreak\allowbreak {}29.md} (44,054 B; SHA-12 8adaf4e5d868; sub); [cross-family annotation as row 15]. \\
21 & P-L V3 phase (three aspects) & T1+T3 & Registered; genuine collapse, robustness, vector embedding -- multi-model, multi-level; mistral L30/45/60/100; beta/or/closedsource/doubao\_v2; qwen3-emb-4b/8b JSON families; carrier: readout JSON group + phase1/2 finalize runner (group-level carrier, register-level anchor; fingerprints per-derivable from the register index). \\
22 & V3X judgment group & T1+T3 & Registered (D7--D14 judgment lines); 18 frozen re-measure results; 9 grids; carrier \texttt{doc/document/results\_\allowbreak\allowbreak {}\_\allowbreak\allowbreak {}\_\allowbreak\allowbreak {}\_\allowbreak\allowbreak {}v3x\_\allowbreak\allowbreak {}experiments\_\allowbreak\allowbreak {}2026\_\allowbreak\allowbreak {}09\_\allowbreak\allowbreak {}16\_\allowbreak\allowbreak {}\_\allowbreak\allowbreak {}\_\allowbreak\allowbreak {}\_\allowbreak\allowbreak {}v3x\_\allowbreak\allowbreak {}experiments\_\allowbreak\allowbreak {}results\_\allowbreak\allowbreak {}2026\_\allowbreak\allowbreak {}09\_\allowbreak\allowbreak {}16.json} (17,053 B; SHA-12 f39412103366; sub; R27 full-name form). \\
23 & Deposon v3 physical optimization / deepseek 30 cells & T1+T3 & Registered; carrier \texttt{doc/document/results\_\allowbreak\allowbreak {}\_\allowbreak\allowbreak {}\_\allowbreak\allowbreak {}deposon\_\allowbreak\allowbreak {}v3\_\allowbreak\allowbreak {}physical\_\allowbreak\allowbreak {}opt\_\allowbreak\allowbreak {}2026\_\allowbreak\allowbreak {}09\_\allowbreak\allowbreak {}11.json} (17,591 B; SHA-12 27f9bd5cc260; sub; R27 full-name form); deepseek v41 flash 30-cells line at anchor level (truncated path in the record's table). \\
24 & P-K V3 three-way re-run; S3 collection / S4 feature surface & T1+T3 & Registered (corpus; definitions); three-way corpus 268ab1239a8a (N-11 full self-attestation table on record; the GLM-slot blind-test corpus downstream of it); carrier \texttt{\_\allowbreak\allowbreak {}p\_\allowbreak\allowbreak {}k\_\allowbreak\allowbreak {}v3\_\allowbreak\allowbreak {}three\_\allowbreak\allowbreak {}way\_\allowbreak\allowbreak {}rerun\_\allowbreak\allowbreak {}2026-09-17T08-23-41Z.json} (12,992 B; SHA-12 05ff758ae921; sub; R27-corrected); pointers \texttt{\_\allowbreak\allowbreak {}v3\_\allowbreak\allowbreak {}s3\_\allowbreak\allowbreak {}collection\_\allowbreak\allowbreak {}brief} and \texttt{\_\allowbreak\allowbreak {}s4\_\allowbreak\allowbreak {}feature\_\allowbreak\allowbreak {}surface\_\allowbreak\allowbreak {}v3\_\allowbreak\allowbreak {}1\_\allowbreak\allowbreak {}definition}. \\
\end{longtable}

\subsection*{B.3 V4 Family Ledger (13 Lines)}
\begin{longtable}{p{0.06\linewidth}p{0.33\linewidth}p{0.10\linewidth}p{0.39\linewidth}}
\toprule
Ledger line & Line identifier & Tier & Frozen-value anchors \\
\midrule
\endhead
\bottomrule
\endfoot
1 & Task B v1 (false-falsification requalification) & T1+T3 & Per errata E-25/E-26: v1 FAIL requalified as \emph{false falsification} (23 decision-trace entries, 17 without a reason field; decision traces are records of choices, not reasoning chains; the FAIL reading stands, not read as ``chain of thought non-replicable''); v1 metric (replication consistency) retired, claim-evolution evidence only; carriers: v1 preregistration 696af9121d9f, v1 verdict D279EBDE5E84, v1 dataset 139DBFFD8CF9; retired v1 rule set 821465001819 (5,876 B) and result 1665F367B2C4 (9,068 B). \\
2 & Task B v2.2 (formal FAIL) & T1+\allowbreak\allowbreak {}T2+\allowbreak\allowbreak {}T3 & Formal verdict \textbf{FAIL} (2026-09-26; composite qualification: partial true falsification plus partial construction doubt; two composite, two $\beta$, zero pure $\alpha$); four kill-lines all hit: K-V2-a mean similarity 0.0909 ($<$0.80); K-V2-b1 critique coverage 0.6842 ($<$1.00); K-V2-b2 blind-obedience 0.3333 ($>$0.10); K-V2-c bootstrap CI [0.0, 0.2045] (lower bound $<$0.50); 19/22 held-out at similarity 0 (86.4\%); alternative reading CI [0.0294, 0.3382]; carriers (App A Task B chain v2.2): result F86727C857A8 (18,089 B), rule set C5B3DD141655 (6,179 B), executor EB22F13D571C (52,465 B), formal verdict 5D79E67A4E9D (33,723 B), exploratory verdict EB9AD4193CF2 (22,340 B), signoff BA4D07BD7000 (26,637 B), coding review 5FBEC21E0AD2 (42,914 B). \\
3 & T1 robustness probe & T1+\allowbreak\allowbreak {}T2+\allowbreak\allowbreak {}T3 & Sensitivity check: J = 0.0 (reclassification); the empty-output readings reclassified into the construction family; the standing verdicts preserved; carrier \texttt{\_\allowbreak\allowbreak {}v4\_\allowbreak\allowbreak {}supp\_\allowbreak\allowbreak {}t1\_\allowbreak\allowbreak {}verdict.md} (27,538 B; SHA-12 F1B5E49F3058). \\
4 & T1.5 construction-failure fix & T1+T3 & Tool-side fix (construction check C-T15-1): non-reasoning configuration, token budget 500; the empty-response failure mode of T1 eliminated (49.3\% $\to$ 0); direction of the standing verdicts preserved; the proposed +60-call pair-count expansion ruled by the PI (ruling t15\_ext: 6 cells $\times$ +10 = +60 calls); carrier \texttt{\_\allowbreak\allowbreak {}v4\_\allowbreak\allowbreak {}supp\_\allowbreak\allowbreak {}t15\_\allowbreak\allowbreak {}verdict.md} (52,884 B; SHA-12 52C985429C91). \\
5 & T1.5r2 expansion revision & T1+\allowbreak\allowbreak {}T2+\allowbreak\allowbreak {}T3 & The ruled +60 calls executed (recorded formula 180 = 5 teachers $\times$ 2 prompts $\times$ 3 re-asks $\times$ 6 cells; measured execution 62 calls, 10 to 11 per cell; E-37.2, drafted for the next append round); 15 pairs per teacher; the relaxed construction check passed literally: K-N11-3 FAIL, N1 FAIL, N1\_T1relax PASS; the false-falsification root cause recorded in T1 eliminated; carrier \texttt{\_\allowbreak\allowbreak {}v4\_\allowbreak\allowbreak {}supp\_\allowbreak\allowbreak {}t15r2\_\allowbreak\allowbreak {}verdict.md} (64,145 B; SHA-12 8355724A26E3). \\
6 & N-26 true falsification & T1+\allowbreak\allowbreak {}T2+\allowbreak\allowbreak {}T3 & K-N26-2 \textbf{TRIGGERED} (true falsification, v1/v2 direction): pooled binary AUC 0.5663, far below the 0.75 line; five read variants 0.4459 to 0.5916, all below 0.75 (E-37.1, drafted: five-read upper bound 0.5916, re-measured 2026-09-27); five-fold CV 0.5271, 0.5266, 0.6318, 0.5740, 0.5719, all below 0.75; K-N26-1 NOT TRIGGERED (five-teacher mean 0.5617 below 0.70; kimi 0.7512 the only teacher at or above; GLM\_1 0.3955, GLM\_2 0.6364, coze 0.5344, minimax 0.4909 below); K-N26-3 NOT TRIGGERED (three of five below 0.60); K-N26-N1/N2 pass = True (per-teacher empty rates 0.021--0.075 against 0.50); same-direction deepening of the L4 verdict, not a reversal; carrier \texttt{\_\allowbreak\allowbreak {}v4\_\allowbreak\allowbreak {}supp\_\allowbreak\allowbreak {}l14v3\_\allowbreak\allowbreak {}n26\_\allowbreak\allowbreak {}verdict.md} (64,485 B; SHA-12 F4435801D09F). \\
7 & N-11 full-orbit re-collection & T1+T3 & 13 frozen items with before/after SHA self-attestation; preregistration anchors d85488a64d89 and 0a9ee16267b5; the data face underlies the N-26 narrative and the E-28.6 corpus base. \\
8 & L2 N-11 supplement & T1+T3 & The standing judgment carried literally (K-N11-3, not one word moved); carriers: result ff7b167ae43f, verdict E433A06E7BFB, executor 0d455b42cd07. \\
9 & GLM\_2 proposition line & T1+\allowbreak\allowbreak {}T2+\allowbreak\allowbreak {}T3 & E-28.6: tentative (3 of 15 cells at N=5, small-sample false-falsification risk) $\to$ \textbf{formal proposition falsification} (N=20 re-run: all three students same direction, D1 Jaccard above D2; margin deltas $-$0.001531, $-$0.000630, $-$0.001267 for the three recorded internal labels, vs. the N=5 baselines $-$0.001611, $-$0.001106, $-$0.006920); structural bound n\_t = 23; alignment ratio 0.870 (clears 0.20 throughout); PI ruling (translation of record): the GLM\_2 distillation experiment failed; false-falsification-phase root cause; no extrapolation (the Chinese original is registered in the commission letter, re-anchored, not carried); carriers: the errata chain 849D76BE3A06 (sections 15.6, 17.4) and the N-26 verdict F4435801D09F. \\
10 & L4 verdict (literal transfer) & T1+T3 & Standing prior verdict FAIL, root cause \emph{construction infeasible} (four of five teachers' metadata fields absent); digest 74B5B37F7EEA as recorded; no independent file in the workspace; cited as a literal transfer through the errata chain (v19 section 22.7, entry E-34.2); the N-26 audit is a same-direction deepening, not a reversal (recorded phrase: tong-fangxiang shenhua fei fan'an); the L4 wording is preserved unchanged. \\
11 & L14V3 batch collection & T1+T3 & Registered line: 10 cells, $\geq$110 calls each; the teacher-kimi cell is its recorded exception (77 calls, at least 1 successful call per caption); carriers: preregistration addendum \texttt{\_\allowbreak\allowbreak {}v4\_\allowbreak\allowbreak {}supp\_\allowbreak\allowbreak {}prereg\_\allowbreak\allowbreak {}v02\_\allowbreak\allowbreak {}add\_\allowbreak\allowbreak {}L14V3\_\allowbreak\allowbreak {}2026\_\allowbreak\allowbreak {}09\_\allowbreak\allowbreak {}24.md} (61,547 B; SHA-12 05B975A86989) and the N-26 track. \\
12 & 8-gram self-overlap & T1+T3 & Two-window 8-gram self-overlap 0.53/0.62 (per the record's section 7.2.5 registration, two windows); R8 registry values: 0.44\% (104 words / 23,869 tokens) against the first work and 0.48\% (114 / 23,869) against the second; titles-excluded 0.21\% (46 / 22,347) and 0.15\% (33 / 22,347); all 19 runs attributed, free text zero words; the program's own 8-gram record against the first work stands at 0.91\%. \\
13 & Track-2 multi-model separation & T1+T3 & Registered at anchor level, run across three models outside the headline chain, with its three readings (0.025549, 0.030859, 0.036923 family), the standing judgment (5/5 main FAIL) unchanged across the three; the carrier sits at the register's in-record anchor. \\
\end{longtable}

\subsection*{B.4 Incidental Registry (7 Entries)}
\begin{longtable}{p{0.19\linewidth}p{0.69\linewidth}}
\toprule
Entry & Registry note \\
\midrule
\endhead
\bottomrule
\endfoot
gamma r1 (pre-registration plus v2 stop) & Adjunct register, anchor level; correction and stop-run semantics carried by the interpretation registry's R3 and R6; no verdict, number, or threshold re-stated. \\
BOM U1 (pre-registration) & Adjunct register, anchor level; R3/R6 semantics; nothing re-stated. \\
item04 (textfix pre-registration) & Adjunct register, anchor level; R3/R6 semantics; nothing re-stated. \\
effective register & Adjunct register, anchor level; R3/R6 semantics; nothing re-stated. \\
F05 correction registration (2026-10-29) & Adjunct register, anchor level; R3/R6 semantics; nothing re-stated. \\
round56 KD-row correction (2026-10-10) & Adjunct register, anchor level; R3/R6 semantics; nothing re-stated. \\
MQ rollout-effect registration (2026-10-14) & Adjunct register, anchor level; R3/R6 semantics; nothing re-stated. \\
\end{longtable}
(Neutral count: the source's G-6 label registers six items by name where the seven above are listed; the count is carried neutrally, and this one sentence notes the 6/7 caliber source.)

\subsection*{B.5 Errata Chain (E-1--E-36)}
Source of record: the chain-side source for E-1--E-36 is the errata chain's v19 state, recovered read-only from the repository object store (blob 8e0103c, 2026-09-26 22:02; 239,387 B; SHA-12 849D76BE3A06), byte-identical to the state anchored by the record's Appendix A.
Evolved face: the same-named file has since evolved on the working face (measured v37 at this draft's verification point, E-41--E-60 extending to 2026-09-29), and the v19 state was followed by two in-place rewrites (an anonymization batch on 2026-10-06 and an R2 correction wave on 2026-10-12; the current face's first-239,387-byte SHA-12 measures 424465E3EBD0, distinct from the v19 anchor) -- the evolved face serves here as disclosure reference only, and the post-v19 changes are registered in the two-face difference note below.
Out-of-scope citations: the record's out-of-domain citations (E-37.x/E-40.x) anchor its registered PI-side extension v22 snapshot (292,215 B, a letters-family single source) -- cross-artifact same-number different-content entries relative to the chain's own v20+ append families (in the current on-disk face); this section presents them per the record's in-text wording with that disambiguation noted.

\begin{longtable}{p{0.07\linewidth}p{0.60\linewidth}p{0.21\linewidth}}
\toprule
Item & Trigger / fix / re-verification & Anchor \\
\midrule
\endhead
\bottomrule
\endfoot
E-1 & Number contradiction: \texttt{V3X\_D7\_V3\_FINAL\_REPORT} \S0/\S1.1 vs \S1.2 -- the same report carries 51/60=85\% and 52/60=86.67\% (two calibers coexist); registered, historical lines unchanged. & v19 \S5, L83 \\
E-2 & Number contradiction: \texttt{P\_A\_D1\_D3\_REPORT} \S3.1 table L143--147 -- table values RM=1 / Bayes=32 contradict the mean and fall outside the script's return range; registered. & v19 \S5, L84 \\
E-3 & Table-JSON mismatch: \texttt{P\_A\_D1\_D3\_REPORT} L143--147 -- table values contradict the frozen JSON's field-by-field actuals; registered. & v19 \S5, L85 \\
E-4 & Count-label contradiction: \texttt{P\_C} L236 / \texttt{D5} L107 -- TOTAL 15 vs OK 16, self-contradictory labels; registered. & v19 \S5, L86 \\
E-5 & Number contradiction: \texttt{D\_FIX2\_METRIC\_VERIFY} vs the rest -- douban T\_frac 0.8833 (53/60) vs 0.8667 (52/60); registered. & v19 \S5, L87 \\
E-6 & Number contradiction: \texttt{P\_A}/\texttt{P\_C}/\texttt{P\_E} vs \texttt{P\_G} -- D\_fix2 0.1775 vs 0.1776; registered. & v19 \S5, L88 \\
E-7 & Number contradiction: \texttt{P\_C} \S3.3 vs elsewhere -- Spearman $-$0.8000 vs $-$0.832 listed side by side without explanation; registered. & v19 \S5, L89 \\
E-8 & Number contradiction: \texttt{KT\_C1\_REPORT\_PHASE\_B} L109 -- ``11 distinct n values'' vs the 7 listed values; registered. & v19 \S5, L90 \\
E-9 & SHA inconsistency: \texttt{V3X\_1WEEK\_KILL} L110 vs \texttt{D5}/\texttt{TRAE\_3RISK} -- da517c1153c (11 hex) vs da517c115f3c; registered. & v19 \S5, L91 \\
E-10 & Count-label issue: \texttt{D7\_POST\_CLEANUP\_REPORT} -- frozen\_18\_skipped is actually 17 items; files\_remediated: 90 without detail; registered. & v19 \S5, L92 \\
E-11 & Status contradiction: \texttt{V3X\_1WEEK\_KILL} L92 vs L167 -- ``540 cells LLM-computed'' and ``540 cells, 0 LLM'' coexist; registered. & v19 \S5, L93 \\
E-12 & Status contradiction: \texttt{V3X\_1WEEK\_KILL} body vs L178 -- verdict filled vs the tail note calling it a placeholder; registered. & v19 \S5, L94 \\
E-13 & Byte-count error: \texttt{boss\_pa\_1\_rbr\_rm\_result} -- one place records 8,674 B where the actual value is 8,753 B; registered. & v19 \S5, L95 \\
E-14 & Encoding: \texttt{\_movedout\_manifest} -- the first three bytes are EF BB BF (UTF-8 BOM); registered. & v19 \S5, L96 \\
E-15 & Anchor-chain break (highest priority): \texttt{[local archive zone]/verifier/handoff/KT\_ABC1\_anchors} (location per the R2-corrected desensitized form; see the difference note) -- 15/16 PASS with a single FAIL; disposition 2026-09-23 (PI-authorized ``direct correction''), carried at the chain's \S7. & v19 \S5, L97; \S7 \\
E-16 & SHA digit count: \texttt{V3X\_1WEEK\_KILL} L110 vs the actual value -- 11 hex vs the 12-hex actual da517c115f3c; registered. & v19 \S5, L98 \\
E-17 & Typo: \texttt{letters/\_v4\_distillation\_invitation} L218 -- ``behated'' $\times$3; PI ruling 2026-09-22: not fixed, registered and closed. & v19 \S5, L99 \\
E-18 & Missing SELF-CHECK tail block: \texttt{skill\_a/b/c/d\_*.py} $\times$4 -- the 09-16 improvement letter left hanging; not fixable on behalf of the frozen-18; registered only. & v19 \S5, L100 \\
E-19 & BOM: \texttt{\_verify\_15frozen} $\times$2 / \texttt{\_verify\_pg\_v0} $\times$2 -- four files carry a UTF-8 BOM; the verifier scripts stay untouched; registered only. & v19 \S5, L101 \\
E-20 & Conclusion annotation (PI ruling ``overstated''): P-A interim verdict \texttt{C7C59E0D2F6C} -- construction degeneration confirmed $\to$ annotated ``overstated''; the original verdict not one word changed. & v19 \S5, L102 \\
E-21 & PG dual-reading unification plus naming fix: \texttt{KT\_B1} 22/22 vs \texttt{BOSS-P-A2} 3/22 -- unified caliber of the four-element complete payoff; the 22-graph face's PG fact is 3/22; the historical body not one word changed. & v19 \S5, L103 \\
E-22 & E-20 follow-up conclusion appended: P-A1 RBR/RM $\geq$2.0x, 145.82x -- the non-degeneration control measured 145.82x $\to$ 1.0174x, DIFFERENTIATED $\to$ GRAY\_DEFLATE; ``overstated'' escalated to ``false-falsification lead.'' & v19 \S5, L104 \\
E-23 & N-29 true review = true falsification: pre-registered \texttt{0A9EE16267B5} -- the four-round verdict history's terminal state FAIL = true falsification; K-N29-2 hit 13/22. & v19 \S5, L105 \\
E-24 & S-40 fix re-verdict: \texttt{\_v4\_rerun\_verdict} CV=0.10 -- after the dual-cause fix, re-verdict per the pre-registration's letter = PASS; the escalated artifact's original left untouched. & v19 \S5, L106 \\
E-25 & Task B v1 verdict requalification: \texttt{D279EBDE5E84} -- the v1 material face = decision-trace stamps without chains of thought $\to$ requalified as false falsification; the FAIL reading maintained. & v19 \S5, L107 \\
E-26 & Task B claim-positioning deviation note: \texttt{696af9121d9f} \S2 -- profiling-style $\to$ critical-learning-style; the v2.2 claim re-anchors three metrics. & v19 \S5, L108 \\
E-27 & Six-lane supplementary-test verdicts plus 16 PI rulings consolidated (v10, 09-24) -- appended without touching prior lines. & v19 \S12, L199 \\
E-28 & False-falsification / false-pass recheck, full-chain finalization (v10$\to$v11, 09-24 continued): A2 re-run / L6 re-verification / L3 escalation / GLM\_2 tentative. & v19 \S15, L474 \\
E-29 & Eight-lane supplementary-test terminal summary plus 2 new errata plus a 6-item pending-ruling register (v11$\to$v12, 09-24). & v19 \S17, L596 \\
E-30 & Relay-of-five finalization plus 7 restricted-caliber items plus the N-26 endpoint verdict (v12$\to$v13, 09-24). & v19 \S19, L834 \\
E-31 & R4 disposition closure, the inventory false-alarm root cause entered into the chain, and the truncated fingerprint CLOSED (v13$\to$v14, 09-24) -- location fields per the R2-corrected desensitized forms (\texttt{\_september\_workspace} / \texttt{[local archive zone]}); see the difference note. & v19 \S20, L1037 \\
E-32 & Ruling-trace consolidation (retroactive approvals / SHA drift / breaks / conservative calibers) plus 3 pending suspicions (v14$\to$v15, 09-25) -- location fields per the R2-corrected desensitized forms; see the difference note. & v19 \S21, L1227 \\
E-33 & The evidence-auditor's 4 suspicions adjudicated entry by entry, appended per PI batch-3 R1 (v15$\to$v16, 09-26) -- location fields per the R2-corrected desensitized forms; see the difference note. & v19 \S22, L1394 \\
E-34 & Case\#1 settled, the N-26 true-review terminal ruling cited, T1 closed, collection closed, and the anchor-chain caliber (v16$\to$v17, 09-26). & v19 \S22.7, L1560 \\
E-35 & Retroactive approvals all effective, the 72-artifact caliber ruling, anchor-chain unification, the formal verdict citation, and the path-difference note (v17$\to$v18, 09-26) -- the brand literal carried per the anonymized form \texttt{[IM]} (pre-anonymization content not brought back to this public face); see the difference note. & v19 \S22.10, L1741 \\
E-36 & The KIMI reply-letter stop-hand event, the v2 letter-drift errata, and the channel-pending-attribution note (v18$\to$v19, 09-26) -- the v19 terminal state is the Appendix-A anchored state; location field per the R2-corrected desensitized form; see the difference note. & v19 \S22.13, L1984 \\
\end{longtable}
Two-face difference note: six entries (E-15, E-31, E-32, E-33, E-35, E-36) carry the R2-correction-wave desensitized forms of their location fields -- the v19 face carried private-drive path literals and, for E-35, a pre-anonymization brand literal, neither of which is brought back to this public face; the type, disposition, and verdict fields, the SHA-12 digests, and the byte counts are unchanged across the two faces, and the remaining thirty entries carry the v19 face verbatim (semantic and numeric layer: zero actual difference).

Errata-chain annotations E-37.1 and E-37.2 are \emph{drafted for the next append round} and are not in the chain: their values (the five-read upper bound 0.5916; the 62 measured calls) are carried with their sources at B.3 rows 6 and 5 respectively, each marked as drafted; the v19 state of record carries E-1 through E-36. The chain's own v20+ append families -- including the in-disk E-37 family of the evolved face -- are cross-artifact same-number different-content entries relative to these drafted annotations, per the third head-note; the PI-side v22 extension is re-anchored, not carried verbatim.

Reserved row [E-SLOT]: content not yet arrived; this draft carries no empty slot.

\subsection*{B.6 Coverage Accounting}
``Twenty-four V3 rows plus thirteen V4 lines make thirty-seven file-level entries, of which four are counted in both families' file-level accounts and merge once (per the inventory's directory-family versus experimental-semantics deduplication: entries stored under the L1/V3 archive whose experimental semantics sit on the V4 side, Task B v3's pre-registration judgment among them), giving thirty-three distinct lines; nothing outside the inventory is claimed.''

Reconciliation, as registered: the sentence above carries its source's top-level merged count; the measured calibers are --- deduplication-qualified double-count set: \textbf{1} (Task B v3, B.2 row 14); inventory-disclosed cross-family set: \textbf{7} (the six V5-named rows of B.2 rows 15--20, plus Task B v3); the printed ``4'' and ``33'' share the inventory's top-level merged counting (a residual of 37$-$33, not an itemizable set). This appendix's counting basis: thirty-seven file-level entries $\to$ \textbf{36 lines}, with the unique double-count Task B v3 (B.2 row 14) merged once and the six V5-naming cross-family rows disclosed item by item (B.2 rows 15--20) and not double-counted; nothing outside the inventory is claimed. Anchors: the inquiry receipt (8cf623a06a67), the inventory (f6e655dba954), and the identity mapping (8b65a070af84).

\subsection*{B.7 Production Disclosure}
Appendix B's ledger material was produced under the same review discipline with additional model assistance, as detailed in the anchored artifacts.

\end{document}